%% file: main.tex
\documentclass[10pt,twocolumn,letterpaper]{article}

\usepackage{iccv}              


\definecolor{iccvblue}{rgb}{0.21,0.49,0.74}
\usepackage[pagebackref,breaklinks,colorlinks,allcolors=iccvblue]{hyperref}

\def\paperID{*****} 
\def\confName{ICCV}
\def\confYear{2025}
\usepackage{graphicx}
\usepackage{booktabs}
\usepackage{amsmath}
\usepackage{subcaption}
\usepackage{booktabs}
\usepackage{float}
\usepackage{stfloats}
\usepackage{mwe}
\usepackage{stfloats}
\usepackage{pgfplots}
\usepackage{pgfplotstable}
\usepackage{xcolor}

\usepackage{hyperref}

\usepackage{tikz}
\usetikzlibrary{shapes.geometric, arrows, calc}
\tikzstyle{startstop} = [rectangle, minimum width=6cm, minimum height=2cm, text centered, draw=black, fill=white]
\tikzstyle{arrow} = [line width=2mm,->,>=stealth]

\title{\LARGE \bf
MR6D: Benchmarking 6D Pose Estimation for Mobile Robots
}

\newcommand{\namesep}{\hspace{0.8em}}

\author{
Anas Gouda$^{1,3}$\namesep
Shrutarv Awasthi$^{1}$\namesep
Christian Blesing$^{2}$\namesep  \\
Lokeshwaran Manohar$^{1}$\namesep
Frank Hoffmann$^{1}$\namesep
Alice Kirchheim $^{1,2,3}$
\vspace{0.56em} \\
  {\normalsize
  $^{1}$TU Dortmund\namesep
  $^{2}$Fraunhofer IML\namesep
  $^{3}$LAMARR Institute for Machine Learning
  }
}%

\begin{document}

\makeatletter
\g@addto@macro\@maketitle{\input{figures/fig_intro}
}
\makeatother

\thispagestyle{empty}
\pagestyle{empty}

\maketitle

\begin{tikzpicture}[remember picture, overlay]
\node[anchor=north, yshift=-4em, text width=0.95\paperwidth, align=center, font=\normalsize\color{gray}] at (current page.north) {
    This paper has been accepted for publication at the\\
    IEEE International Conference on Computer Vision (ICCV) Workshops, Honolulu, 2025. ©IEEE
};
\end{tikzpicture}

\input{sec/0_abstract}    
\input{sec/1_intro}
\input{sec/2_related_work}
\input{sec/3_collection}
\input{sec/4_evaluation}

\input{sec/5_conclusion_future_work}

{
    \small
    \bibliographystyle{ieeenat_fullname}
    \bibliography{references}
}


\end{document}

%% file: figures/fig_intro.tex
\begin{figure}[H]
\vspace{-20pt}
    \centering
    \begin{minipage}{\textwidth} 
        \centering
        \subfloat[O³dyn \cite{o3dyn}]{%
            \includegraphics[height=3.55cm, trim=100 0 100 0, clip]{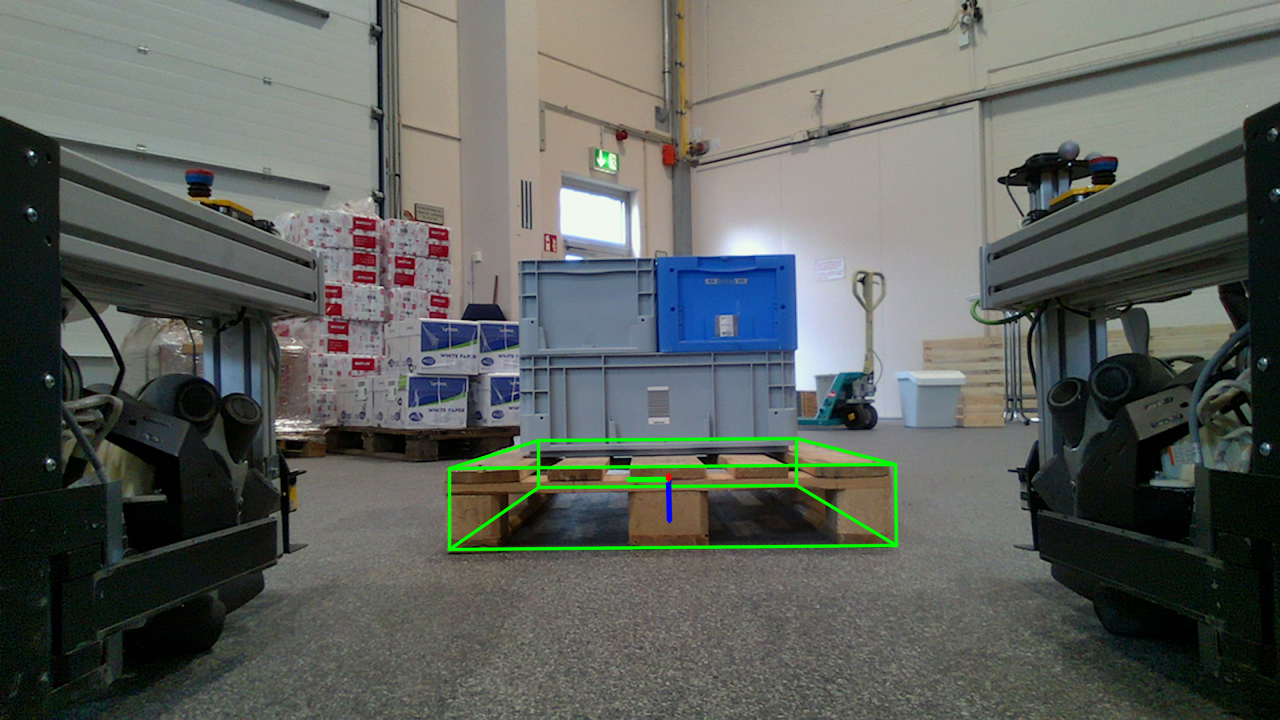}
            \hspace{-0.3cm}
            \includegraphics[height=3.55cm]{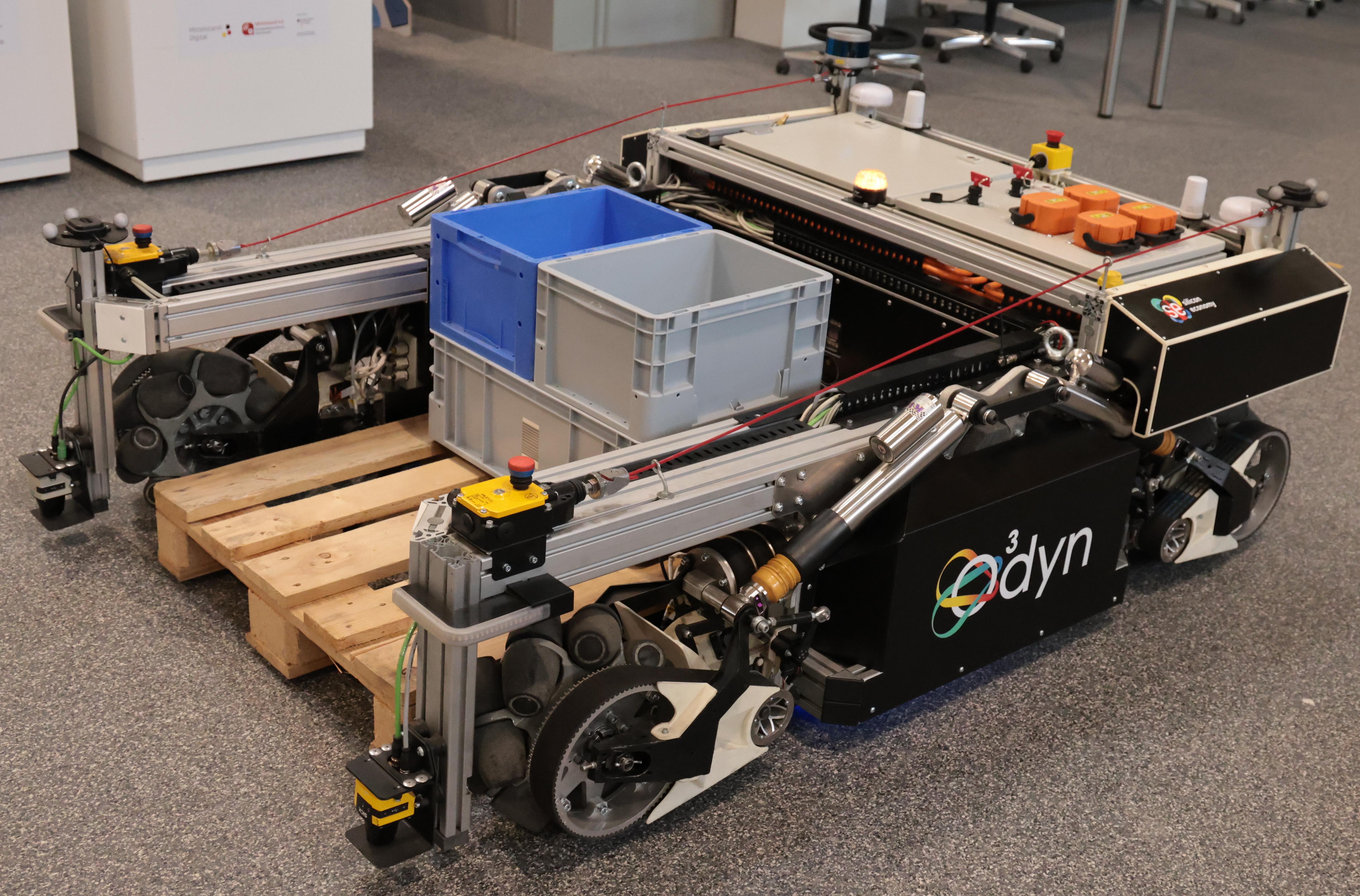}
        }
        \hfill
        \subfloat[evoBot \cite{evobot}]{%
            \includegraphics[height=3.55cm, trim=600 50 0 50, clip]{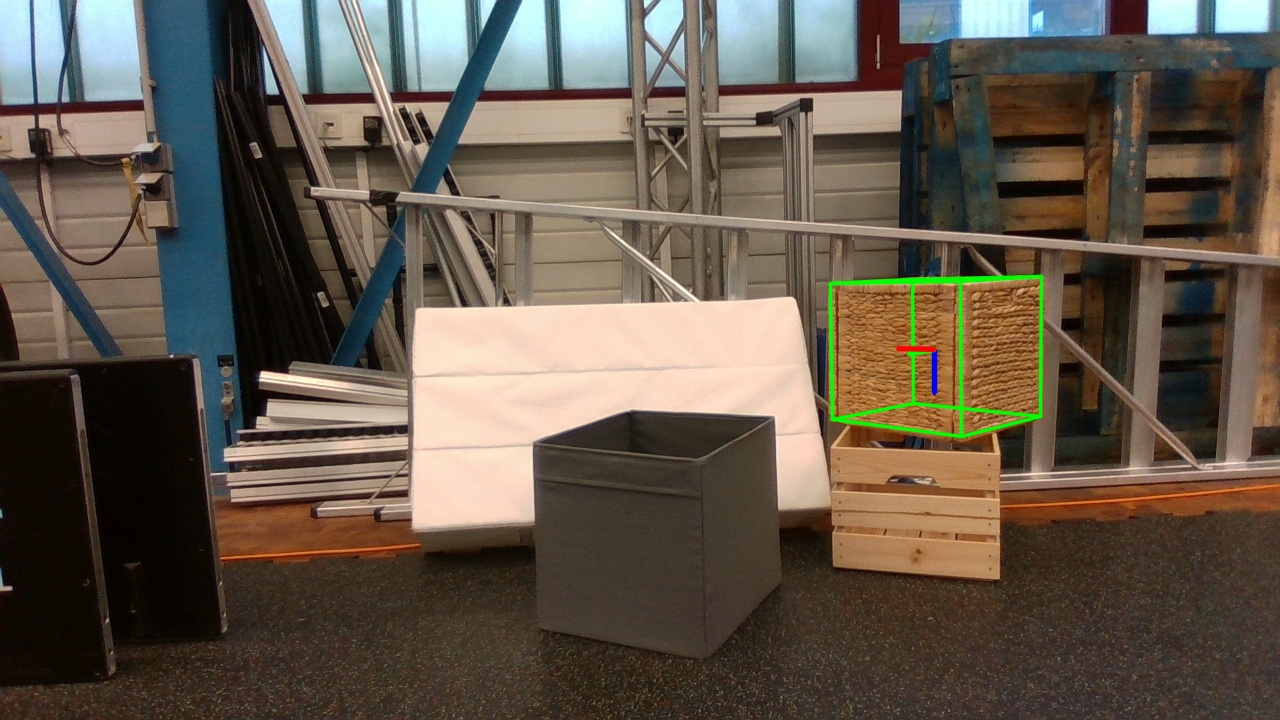}
            \hspace{-0.3cm}
            \reflectbox{\includegraphics[height=3.55cm]{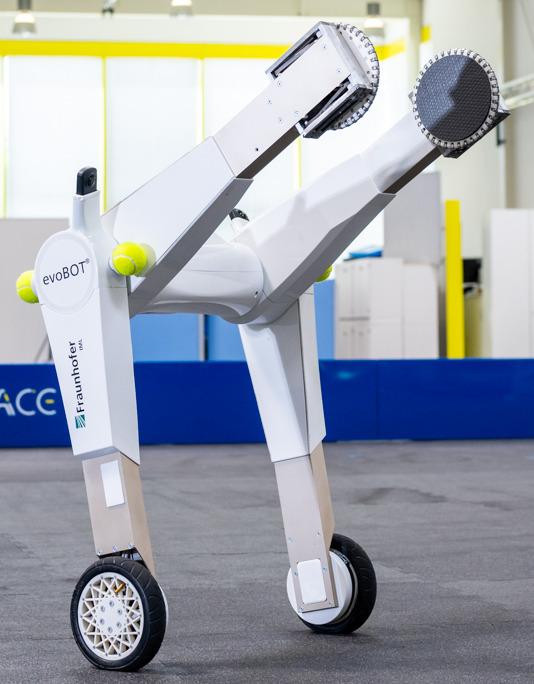}}
        }
    \end{minipage}

    \vspace{0.2cm} 

    \captionsetup{width=\textwidth}  
    \makebox[\textwidth]{  
        \parbox{0.95\textwidth}{ 
            \setcounter{figure}{0} 
            \captionof{figure}{Many mobile robot platforms use specialized gripping mechanisms rather than standard robotic arms, allowing them to handle larger objects positioned farther from the robot’s camera, as demonstrated by the open-source robots O³dyn and evoBOT. Unlike most 6D pose datasets focused on household objects, MR6D highlights the unique challenges mobile robots face, including long-range detection, extreme camera perspectives and self-occlusions. The figure showcases an example of such robots, featuring real camera perspectives from our dataset for the O³dyn and a similar perspective to the evoBOT.}

            \label{fig:intro}
        }
    }
\end{figure}

%% file: sec/0_abstract.tex
\begin{abstract}

\vspace{-30pt}

Existing 6D pose estimation datasets primarily focus on small household objects typically handled by robot arm manipulators, limiting their relevance to mobile robotics. Mobile platforms often operate without manipulators, interact with larger objects, and face challenges such as long-range perception, heavy self-occlusion, and diverse camera perspectives. While recent models generalize well to unseen objects, evaluations remain confined to household-like settings that overlook these factors. We introduce MR6D, a dataset designed for 6D pose estimation for mobile robots in industrial environments. It includes 92 real-world scenes featuring 16 unique objects across static and dynamic interactions. MR6D captures the challenges specific to mobile platforms, including distant viewpoints, varied object configurations, larger object sizes, and complex occlusion/self-occlusion patterns. Initial experiments reveal that current 6D pipelines underperform in these settings, with 2D segmentation being another hurdle. MR6D establishes a foundation for developing and evaluating pose estimation methods tailored to the demands of mobile robotics. The dataset is available at \href{https://huggingface.co/datasets/anas-gouda/mr6d}{https://huggingface.co/datasets/anas-gouda/mr6d}. 

\input{figures/fig_datasets_comparison}
\end{abstract}

%% file: figures/fig_datasets_comparison.tex
\begin{figure*}[t]
    \centering
    \begin{tikzpicture}
        \begin{axis}[
            xlabel={Average Distance from Camera (mm)},
            ylabel={Average Object Volume (cm$^3$)},
            title={Dataset Comparison: Distance vs. Average Object Volume},
            grid=major,
            width=9cm, 
            height=5cm, 
            xmin=0, ymin=0, 
            legend style={
                at={(1.05,0.5)}, 
                anchor=west,
                legend columns=3, 
                font=\small, 
                row sep=0.1cm 
            }
        ]

        \addplot[color=red, only marks, mark=*] coordinates {(550.04, 1.66)};
        \addplot[color=blue, only marks, mark=*] coordinates {(713.93, 0.41)};
        \addplot[color=green, only marks, mark=*] coordinates {(1002.71, 1.83)};
        \addplot[color=orange, only marks, mark=*] coordinates {(876.95, 2.32)};
        \addplot[color=purple, only marks, mark=*] coordinates {(768.84, 0.35)};
        \addplot[color=brown, only marks, mark=*] coordinates {(734.24, 0.25)};
        \addplot[color=cyan, only marks, mark=*] coordinates {(888.55, 2.60)};
        \addplot[color=magenta, only marks, mark=*] coordinates {(883.74, 1.19)};
        \addplot[color=teal, only marks, mark=*] coordinates {(749.25, 1.12)};
        \addplot[color=pink, only marks, mark=*] coordinates {(1148.23, 7.64)};
        \addplot[color=lime, only marks, mark=*] coordinates {(648.62, 0.92)};
        \addplot[color=violet, only marks, mark=*] coordinates {(827.91, 1.13)};
        \addplot[color=gray, only marks, mark=*] coordinates {(779.07, 2.10)};

        \addplot[color=black, only marks, mark=*] coordinates {(2078.28, 57.45)};
        \addplot[color=black!70, only marks, mark=triangle*] coordinates {(1745, 57.45)}; 
        \addplot[color=black!60, only marks, mark=triangle*] coordinates {(2130.68, 57.45)};
        \addplot[color=black!50, only marks, mark=triangle*] coordinates {(2549.79, 57.45)};
        \addplot[color=black!40, only marks, mark=triangle*] coordinates {(1715, 57.45)}; 

        \addlegendimage{mark=*, color=red} \addlegendentry{HOT-3D}
        \addlegendimage{mark=*, color=blue} \addlegendentry{HOPE}
        \addlegendimage{mark=*, color=green} \addlegendentry{HANDAL}
        \addlegendimage{mark=*, color=orange} \addlegendentry{Linemod}
        \addlegendimage{mark=*, color=purple} \addlegendentry{T-LESS}
        \addlegendimage{mark=*, color=brown} \addlegendentry{ITODD}

        \addlegendimage{mark=*, color=cyan} \addlegendentry{HomebrewedDB}
        \addlegendimage{mark=*, color=magenta} \addlegendentry{YCB-V}
        \addlegendimage{mark=*, color=teal} \addlegendentry{IC-BIN}
        \addlegendimage{mark=*, color=pink} \addlegendentry{TUD-L}
        \addlegendimage{mark=*, color=lime} \addlegendentry{RU-APC}
        \addlegendimage{mark=*, color=violet} \addlegendentry{TYO-L}

        \addlegendimage{mark=*, color=gray} \addlegendentry{ICMI}
        \addlegendimage{mark=*, color=black} \addlegendentry{MR6D (All)}
        \addlegendimage{mark=triangle*, color=black!70} \addlegendentry{MR6D (Val)}
        \addlegendimage{mark=triangle*, color=black!60} \addlegendentry{MR6D (Dynamic)}
        \addlegendimage{mark=triangle*, color=black!50} \addlegendentry{MR6D (O³dyn)}
        \addlegendimage{mark=triangle*, color=black!40} \addlegendentry{MR6D (MR)}

        \end{axis}
    \end{tikzpicture}
    \caption{\textbf{Dataset Comparison:} Average object distance from the camera and average object volume. Triangles indicate MR6D subsets. "MR6D (all)" represents the mean across its subsets, rather than the global per-scene average. For other datasets, averages are computed over the validation or test subset, depending on public availability.}
    \label{fig:dataset_comparison}
\end{figure*}
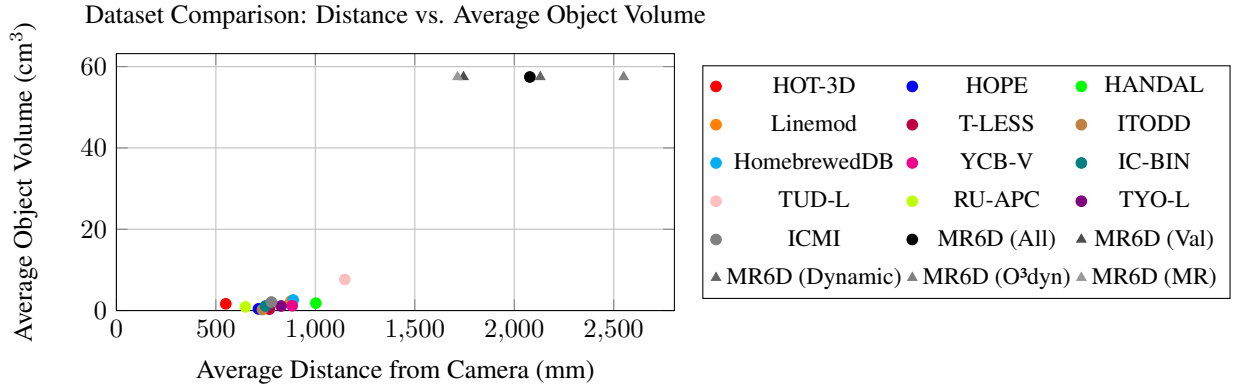

%% file: sec/1_intro.tex
\vspace{-3mm}

\section{Introduction}


6D pose estimation is crucial for robotic perception, enabling accurate object localization and manipulation. Existing datasets such as \cite{handal, hot3d, hope, lm, tless, itodd, hb, ycb-v} have advanced pose estimation research but mainly focus on small household objects typically manipulated by robotic arms, limiting their applicability to mobile robots. Unlike robotic arms, industrial mobile robots rely on specialized gripping mechanisms tailored to their specific tasks. Figure~\ref{fig:intro} illustrates two open-source mobile robots: O³dyn, an omnidirectional robot for high-speed pallet transport, and evoBOT, a self-balancing robot for warehouse automation. These two platforms exemplify the diverse motion patterns found in industrial mobile robotics---from high-speed, omnidirectional navigation to agile, self-balancing maneuvers---providing a representative view of how mobile robots operate in real-world settings. Their operation highlights the critical role of precise 6D pose estimation in real-world mobile robotics.


Mobile robot perception introduces challenges not present in robotic arm setups. While robotic arms observe objects from a fixed downward perspective, mobile robots encounter diverse camera perspectives—from looking up at objects to seeing them at eye level with limited visibility. They must detect larger objects at greater distances while handling occlusions and extreme viewpoints, increasing pose estimation difficulty. Sparse or missing depth data at long range further complicates object recognition and pose estimation.


Mobile robots in industry require instance-level pose estimation. Industrial robots often perform repetitive tasks where speed and deterministic execution are prioritized over generalization. Unlike applications requiring broad generalization across object categories, industrial robots typically handle specific known objects. For example, a robot may be programmed to manipulate a specific type (instance) of pallets, such as Euro pallets, rather than handling all possible pallet designs. The handling process is usually structured and optimized for predefined objects. If the pallet type changes, adjustments in the process or task execution are usually required rather than relying purely on a generalized pose estimation model. Therefore, we create our dataset with instance-level pose estimation instead of category-level.


To address these challenges, we introduce a new dataset specifically designed for mobile robots. Unlike existing datasets focused on small household objects, ours captures larger objects in varied environments. It includes both static and dynamic scenes, where object visibility, occlusion, and perspective change due to robot movement or external factors. These aspects—long-range detection, non-standard viewpoints, and diverse grasping mechanisms—are underrepresented in existing datasets. By incorporating them, our dataset provides a more realistic benchmark for mobile robot perception.
The dataset focuses on objects relevant to mobile robots, including pallets, storage bins, and widely available consumer storage items. It also includes multiple instances of the same object, closely stacked configurations, and objects with similar colors and textures to test the robustness of detection and segmentation pipelines.


We evaluate our dataset using two setups: one combining ground-truth 2D masks with 6D pose estimation of unseen objects, and another using a fully unseen pipeline where both 2D segmentation and 6D pose estimation are unseen. Results indicate that while the 6D pose estimation component shows some generalization to unseen objects, overall performance is influenced by segmentation quality.


In summary, MR6D targets critical perception challenges in mobile robot perception and provides a realistic benchmark to advance 6D pose estimation of unseen objects in industrial environments.
The remainder of this paper is structured as follows: Section~\ref{sec:related_work} reviews existing datasets and approaches, Section~\ref{sec:dataset} outlines the dataset collection process, and Section~\ref{sec:evaluation} presents benchmarking results and performance limitations.

%% file: sec/2_related_work.tex
\section{Related Work}
\label{sec:related_work}

6D pose estimation can be categorized into instance-level and category-level approaches. Instance-level pose estimation targets specific, known objects, while category-level methods generalize to unseen instances within a category. This work focuses on instance-level pose estimation, which is particularly relevant for industrial and mobile robotics applications, where explicit processes rely on specific object models.


The BOP Challenge is a standardized benchmark for 6D pose estimation, compiling datasets designed to evaluate object pose estimation methods. It includes widely used datasets such as LM \cite{lm}, T-LESS \cite{tless}, HB \cite{hb}, YCB-V \cite{ycb-v}, IC-BIN \cite{icbin}, and TUD-L \cite{bop18}. These datasets encompass various object types commonly used in benchmarking 6D pose estimation models. While ITODD \cite{itodd} focuses on industrial bin picking, most datasets primarily feature small household objects or interactions centered around robotic arms. HOT3D, for example, targets egocentric hand-object interactions for Augmented Reality (AR) applications, further highlighting the household-centric bias. Although these datasets have advanced 6D pose estimation, their focus on small household objects limits their relevance to mobile robotics, where objects are often larger, require non-standard handling, or are not manipulated by robotic arms at all. MR6D addresses these gaps by introducing larger objects and capturing challenges unique to mobile robotics, such as long-range detection, diverse camera perspectives, and occlusions, making it a more suitable benchmark for real-world applications.

To further illustrate our dataset’s relevance for mobile robotics, we compare it against existing datasets in terms of average object volume and object distance from the camera. Figure \ref{fig:dataset_comparison} shows that MR6D contains larger objects and greater detection distances, aligning with real-world mobile robot perception challenges.


Category-level datasets aim to generalize 6D pose estimation across object categories rather than specific instances. These datasets typically contain a diverse range of object classes, allowing models to learn category-wide shape and pose variations. Several category-level datasets have been developed, such as Objectron, which focuses on everyday objects with annotated 3D bounding boxes. The HANDAL dataset covers both instance-level and category-level pose estimation, offering a large-scale collection of annotated frames with detailed 6D pose, scale information, and affordance labels, specifically tailored for robotic manipulation. While category-level pose estimation is valuable for general-purpose applications, it is less applicable in industrial and mobile robotics settings, where known object instances are explicitly modeled and manipulated. In such environments, precise instance-level pose estimation remains the priority.


The aforementioned 6D pose estimation datasets employ a range of camera tracking methods for annotation, including fiducial markers, scene reconstruction via NeRF or Gaussian Splatting, and motion capture systems. Some approaches estimate camera poses from multi-view images, while others rely on real-time tracking using motion capture or structure-from-motion (SfM). In static scenes, annotations are typically based on precomputed camera trajectories, whereas dynamic scenes require simultaneous tracking of both the objects and the camera. For our dataset, we adopt multiple annotation strategies across different subsets. The dynamic subset and one of the static subsets are annotated using a motion capture system. For the remaining subsets, we use SfM, robot odometry, and fiducial markers.


6D pose estimation methods can be broadly divided into seen-object and unseen-object approaches. Seen-object methods are trained on specific objects and rely on detailed models to infer 6D poses accurately. Examples include CosyPose \cite{cosypose} and GDRNPP \cite{gdrnpp}, which employ deep learning-based refinement techniques for precise pose estimation of known objects.

Unseen-object methods generalize to novel objects without requiring prior training data. These approaches typically follow a pipeline consisting of segmentation, identification, and 6D pose estimation. Recent methods such as MegaPose \cite{megapose}, GigaPose \cite{gigaPose}, FoundPose \cite{foundpose}, FoundationPose \cite{foundationpose}, and GenFlow \cite{genflow} utilize strategies including retrieval-based estimation, generative flow models, and large-scale pretraining to enhance generalization. As these methods do not include object segmentation, they depend on separate segmentation pipelines such as CNOS \cite{cnos} or CTL \cite{ctl} for object identification. For our evaluation, we use SegmentAnything2 \cite{sam2} for classification, CTL \cite{ctl} for object identification, and FoundationPose \cite{foundationpose} for 6D pose estimation.


MR6D complements existing benchmarks by focusing on industrially relevant objects and perception challenges specific to mobile robotics, such as long-range detection and dynamic interactions. Its instance-level annotations support precise, task-driven pose estimation, which is critical in structured, real-world environments.

%% file: sec/3_collection.tex
\section{Data Collection}
\label{sec:dataset}

In this section, we describe the dataset collection process, starting with the selection criteria and the creation of 3D mesh models for the dataset objects. We then detail the data collection methodology for the four subsets of MR6D: Validation Static, Dynamic Test, O³dyn Static Test, and Mobile Robot-Like (MR) Static. These subsets include both static and dynamic scenes, captured to reflect diverse and realistic mobile robotic perception scenarios. Figure~\ref{fig:dataset_subsets} shows sample frames from each subset, highlighting the variety of environments and object configurations represented in the dataset.

\subsection{Objects and Modeling}

\input{figures/fig_mr6d_objects}

The selection of objects in MR6D follows two main criteria: (1) ensuring that the objects cannot be grasped using a standard two-finger or suction gripper, and (2) prioritizing standardized and globally available objects. Figure~\ref{fig:mr6d_objects} presents the 16 objects included in our dataset.

Our dataset features a Euro pallet, a widely used industrial standard. Additionally, it includes three Euro standard storage containers (KLT bins) of varying sizes and colors, two of which are identically colored, introducing challenges in object differentiation. We also incorporate an Amazon Basics suitcase and 11 storage items from IKEA, ensuring that all selected IKEA objects will remain available through at least the end of 2026, to support reproducibility and future research.

For all objects, we provide high-quality 3D mesh models. The Euro pallet mesh is manually designed, while the remaining objects are reconstructed using BundleSDF \cite{bundlesdf}. The RGB and depth images for BundleSDF reconstruction are captured using a Zivid 2 camera, which provides highly accurate depth measurements. This accuracy is critical for achieving high-quality reconstructions, especially compared to those produced with RealSense cameras. Since our dataset does not include a dedicated training set, these object meshes can be used to generate synthetic training data for seen-object 6D pose estimation models, similar to other datasets in the BOP benchmark that utilize BlenderProc \cite{blenderproc}.

\subsection{Collection Environment and Process}

\begin{figure}
\centering
    \includegraphics[height=6.3cm, trim= 220 45 270 140, clip]{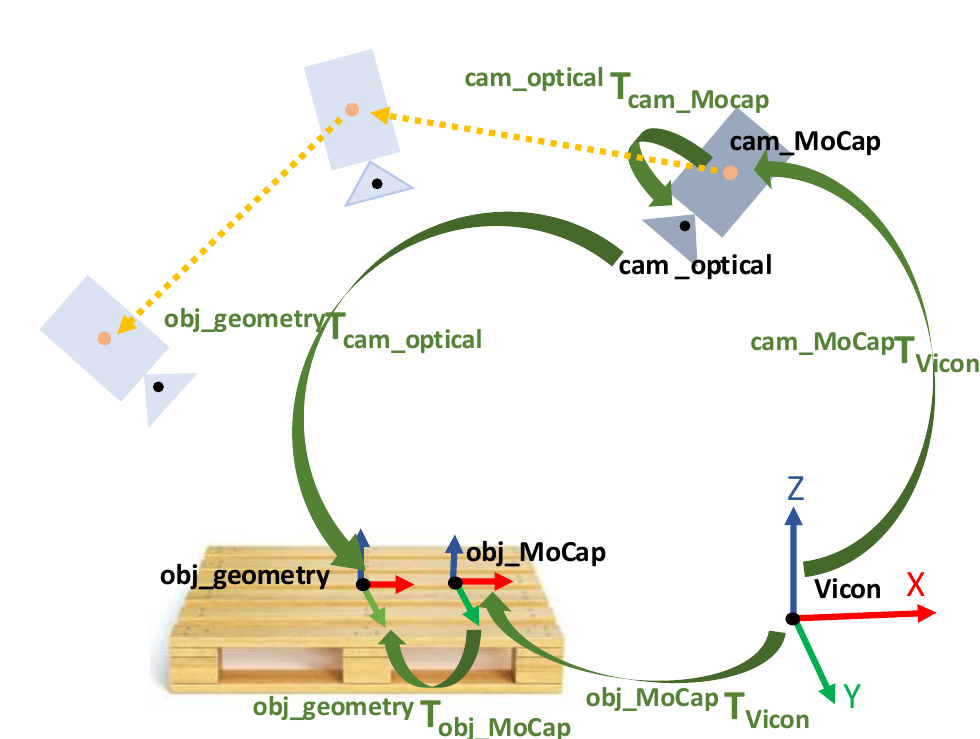}
    \caption{
    The figure illustrates the coordinate transformations used in the validation and dynamic subsets. The world frame corresponds to the origin of the Vicon motion capture (MoCap) system. Two calibrations are performed: (1) Eye-in-hand calibration to align the MoCap-tracked camera with its optical frame, and (2) Object calibration, used in the dynamic subset to align the MoCap-tracked object with its geometric reference frame. These calibrations eliminate alignment errors and ensure accurate 6D pose annotations. In the validation subset, where objects are static, the transformation between the \texttt{Vicon} world and \texttt{obj\_MoCap} is measured only once, as illustrated in Figure~\ref{fig:manual_anno}. The other static subsets (O³dyn and MR-Like) do not require these calibrations.
    }
    \label{fig:transformations}
\end{figure}

\input{figures/fig_manual_anno}
\input{figures/fig_collection_process}
\input{figures/fig_data_statistics}

MR6D was collected across multiple environments using different camera setups tailored to each subset.

\textbf{Validation subset.} This subset is captured in a research hall (hangar building) with a 22×10m² robot arena equipped with a VICON motion capture system consisting of 55 cameras. The data is collected at 30 FPS using a RealSense D435i camera. The camera is tracked using the VICON system. Since the motion capture tracking provides an arbitrary frame on the camera, an eye-in-hand calibration is carried out between the VICON-tracked frame and the camera's optical frame using TSAI method \cite{tsai_handeyecalib}, as shown in Figure~\ref{fig:transformations} (\texttt{cam\_MoCap} to \texttt{cam\_optical}). Objects in this subset are static. Figure~\ref{fig:collection_process} shows the collection process. To capture a scene, objects are placed and their transformations are recorded using a marker object, as shown in Figure~\ref{fig:manual_anno}. This initial marking is not accurate and is manually refined after data collection. This step is necessary because the objects are not tracked by the motion capture system, in order to avoid having large VICON markers visible in close-up images. Only the camera is tracked by VICON. Scenes are captured by moving the camera along a half-circular path around the object. After capturing a scene, a global point cloud is assembled using the camera trajectory, and the manual refinement of object positions is performed using the annotation tool from the BOP toolkit~\cite{bop2018}. This makes the collection process for this subset nearly fully automated, with the only manual step being the final refinement. The last step projects per-frame annotations using the known camera trajectory.

\textbf{Dynamic subset.} This subset is captured in a research hall similar to the one used for the validation subset, but with a larger robot arena measuring 30~×~15~m². This subset is collected at 30~FPS using an Intel RealSense D435i camera. The scenes include a human interacting with the object (moving, transporting, picking, and stowing), simulating a collaborative setting in which a robot picks an object directly from a person. This is the only subset in which both the camera and objects are in motion. The average distance between the camera and the object is maintained near the maximum range of the RealSense camera. Objects, each equipped with motion capture markers, are tracked using the VICON motion capture system, making the data collection process fully automated. To prevent these markers from being exploited for 6D pose extraction, their placement is changed several times for the same object. The same eye-in-hand calibration as in the validation subset is performed. As shown in Figure~\ref{fig:transformations}, an additional calibration---similar to the camera calibration---is required between the VICON-tracked object frame (\texttt{obj\_MoCap}) and the object’s geometric center defined in the mesh model (\texttt{obj\_geometry}). To perform this, we captured multiple frames of each object with VICON tracking, manually aligned the poses using the BOP annotation tool, recorded the offsets, and computed calibration values per object. These two calibrations ensure accurate tracking and eliminate alignment errors.

\textbf{O³dyn subset.} This subset is captured at 30~FPS using the robot’s built-in Intel RealSense L515 LiDAR camera, mounted 40~cm above the ground to provide a distinct low-angle perspective. Scenes are recorded both indoors and outdoors under varying lighting conditions, including direct sunlight and shadows. It is the most challenging subset due to large detection distances and severe depth degradation in sunlight, often resulting in nearly empty depth images. Data is collected in industrial-like environments, such as a small warehouse prototype and a workshop floor. No external tracking of the robot’s camera is available. The initial camera trajectory is estimated from the robot’s odometry. After capture, we recover the camera trajectory and a sparse point cloud using the VGGT model~\cite{vggt}, a feed-forward transformer for multi-view 3D reconstruction. Because VGGT, like traditional SfM, does not recover absolute scale, the odometry-based trajectory is used to compute a scale factor. In scenes with noisy odometry (e.g., due to wheel slip), scale is manually refined with an extended BOP manual annotation tool, aligning the point cloud to real-world dimensions using known object models. Once scaled, 6D object poses are annotated manually, and the refined camera trajectory together with these poses is used to generate per-frame 6D annotations for all objects in the scene.

\textbf{MR-like subset.} This subset is captured at 30 FPS using the Intel RealSense D435i camera. The collection process is similar to that of the O³dyn subset, but the scaling is performed entirely manually using the BOP manual annotation tool. This subset simulates the movement of a wheeled robot, such as the evoBOT. The simulated motion starts from a far distance, approaches the objects, and ends with the camera focusing on one of the objects—mimicking a robot preparing to grasp it.

\textbf{Quality of the annotations.} We track the camera and objects using Vicon motion capture or VGGT \cite{vggt}. Large object distances in some scenes make tracking more challenging and can amplify small errors. Static subsets generally have high accuracy due to stable poses and fixed placement, while dynamic subsets may show slightly larger deviations because both the camera and objects are tracked simultaneously and markers can be occluded. Certain objects (IDs~9, 10, and 15) are not completely rigid and may deform slightly. Additionally, Euro pallets in general are not manufactured to exact nominal dimensions and can vary by a few centimeters. Based on our annotation process and visual checks, we expect static scenes to be accurate within the low-centimeter range.



Figure~\ref{fig:data_statistics} presents key statistics of our dataset. The size of MR6D is comparable to the test subsets used in the BOP benchmark.

\input{figures/fig_dataset_subsets}

%% file: figures/fig_mr6d_objects.tex
\begin{figure}

    \centering
    \begin{minipage}{0.175\textwidth} 
        \centering        \includegraphics[width=2.25cm,height=2.25cm,keepaspectratio]{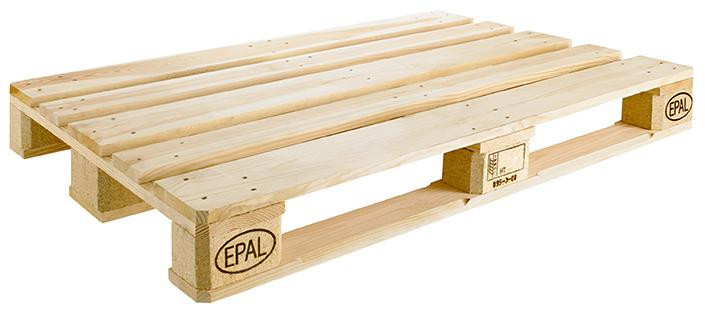}        
    \end{minipage}    
    \begin{minipage}{0.06\textwidth}
        \centering        \includegraphics[width=\textwidth,height=1cm,keepaspectratio]{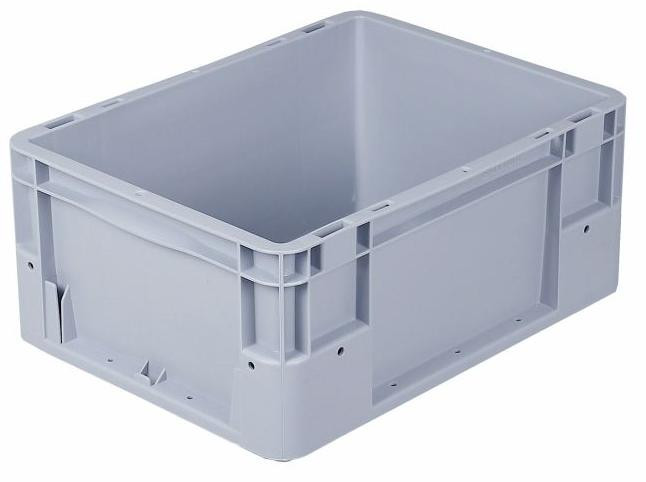}
    \end{minipage}    
    \begin{minipage}{0.075\textwidth}
        \centering        \includegraphics[width=\textwidth,height=1cm,keepaspectratio]{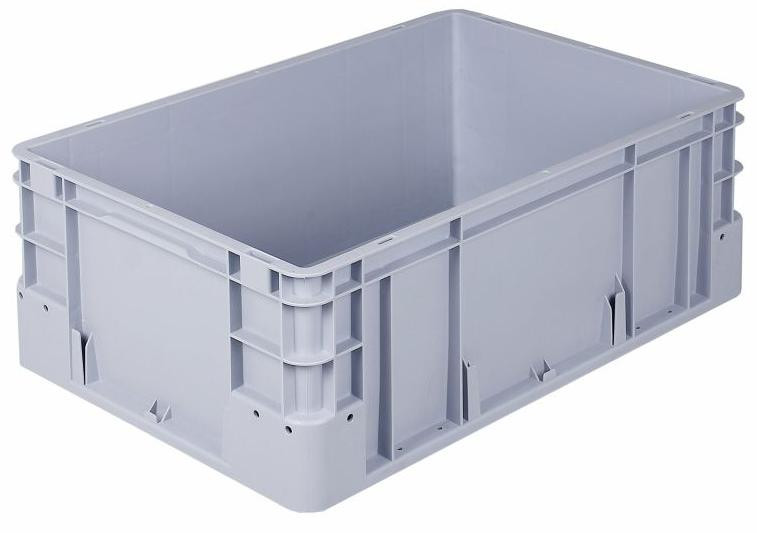}
        
    \end{minipage}    
    \begin{minipage}{0.055\textwidth}
        \centering        \reflectbox{\includegraphics[width=\textwidth,height=1cm,keepaspectratio]{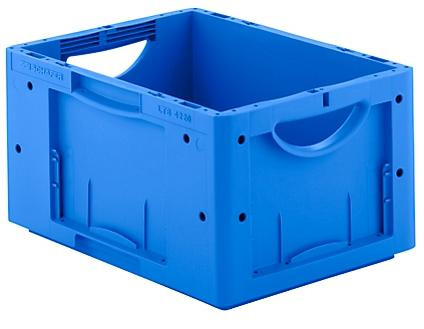}}
        
    \end{minipage}    
    \begin{minipage}{0.07\textwidth}
        \centering        \includegraphics[width=\textwidth,height=1cm,keepaspectratio]{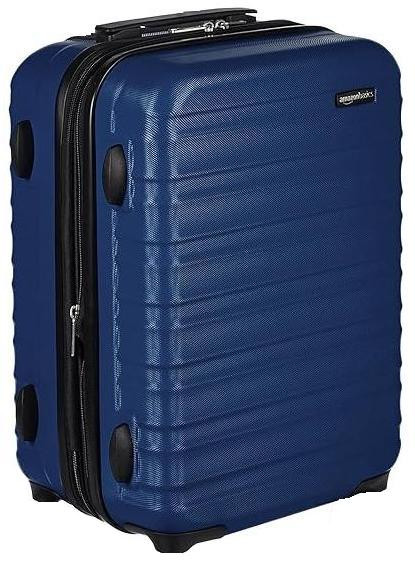}
    \end{minipage}

    \begin{minipage}{0.05\textwidth}
        \centering      
        \reflectbox{\includegraphics[width=1cm,height=1cm]{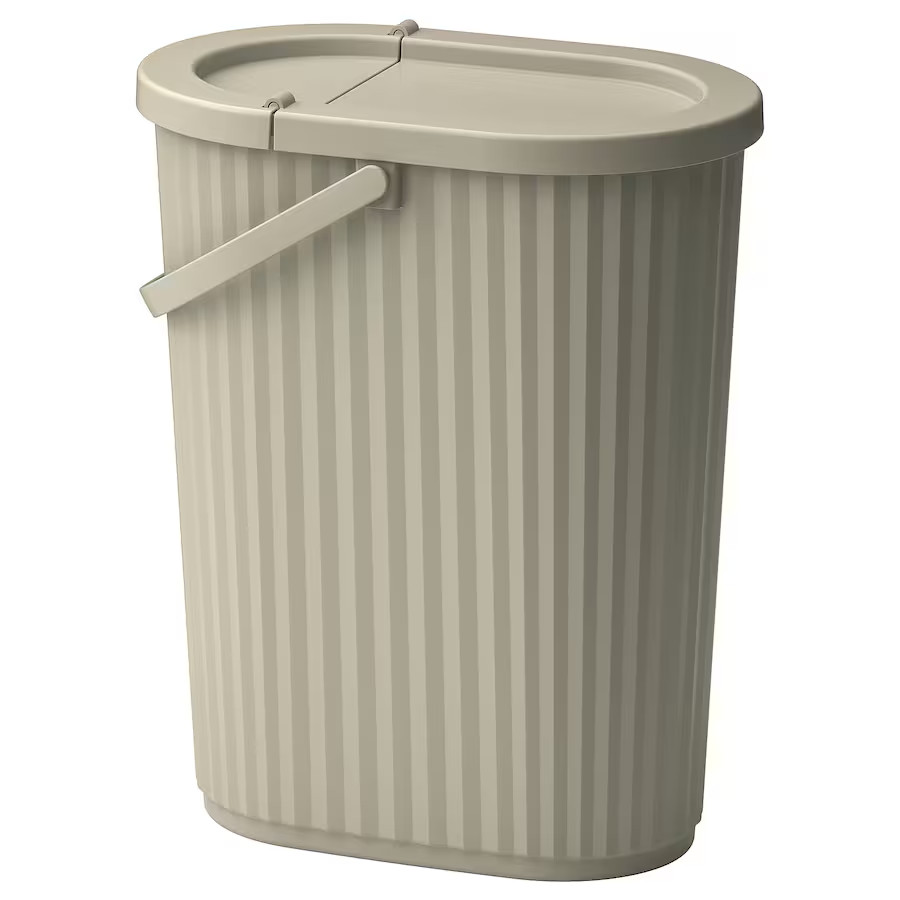}}
    \end{minipage}
    \begin{minipage}{0.07\textwidth}
        \centering        \reflectbox{\includegraphics[width=1.7cm,height=1.7cm,keepaspectratio]{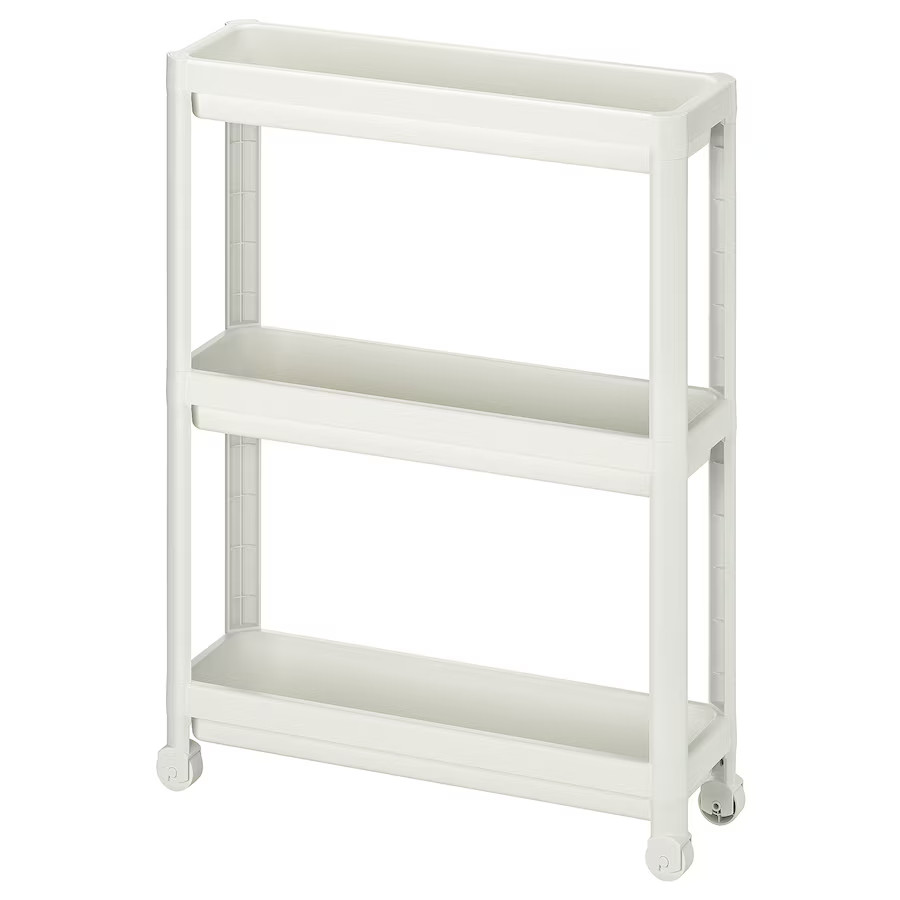}}
    \end{minipage}
    \begin{minipage}{0.09\textwidth}
        \centering
        \includegraphics[width=1.55cm,height=1.3cm,keepaspectratio]{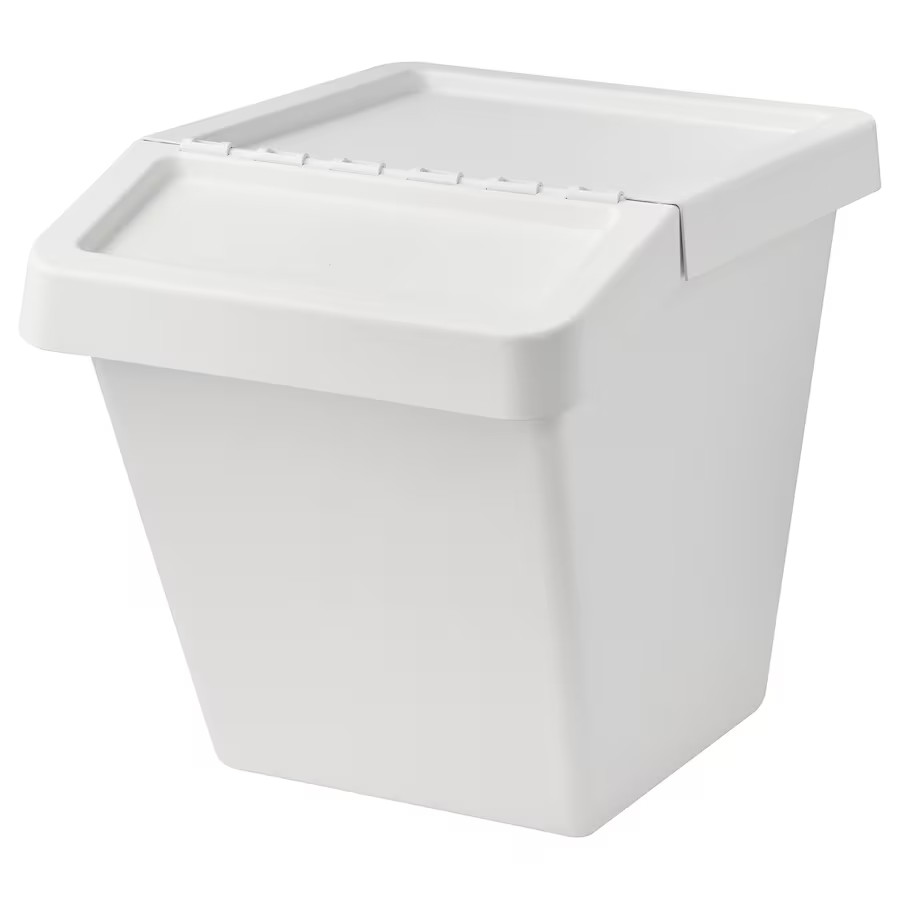}
    \end{minipage}
    \begin{minipage}{0.06\textwidth}
        \centering        \includegraphics[width=\textwidth,height=1cm,keepaspectratio]{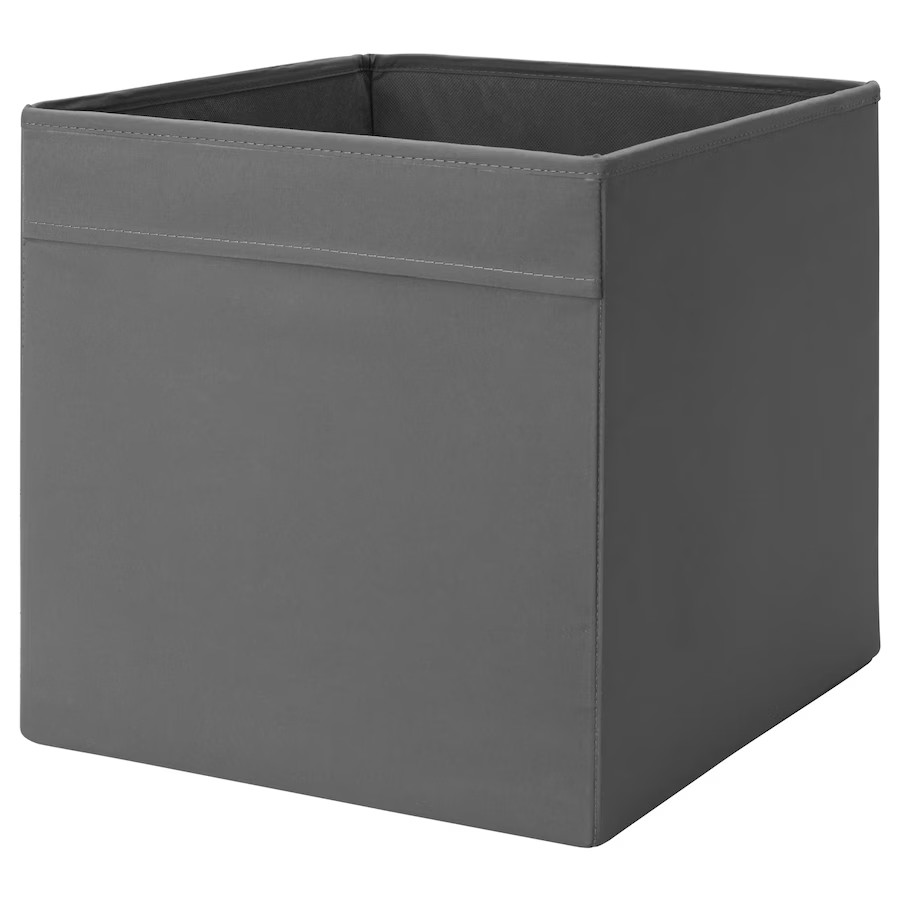}
    \end{minipage}
    \begin{minipage}{0.06\textwidth}
        \centering        \includegraphics[width=\textwidth,height=1cm,keepaspectratio]{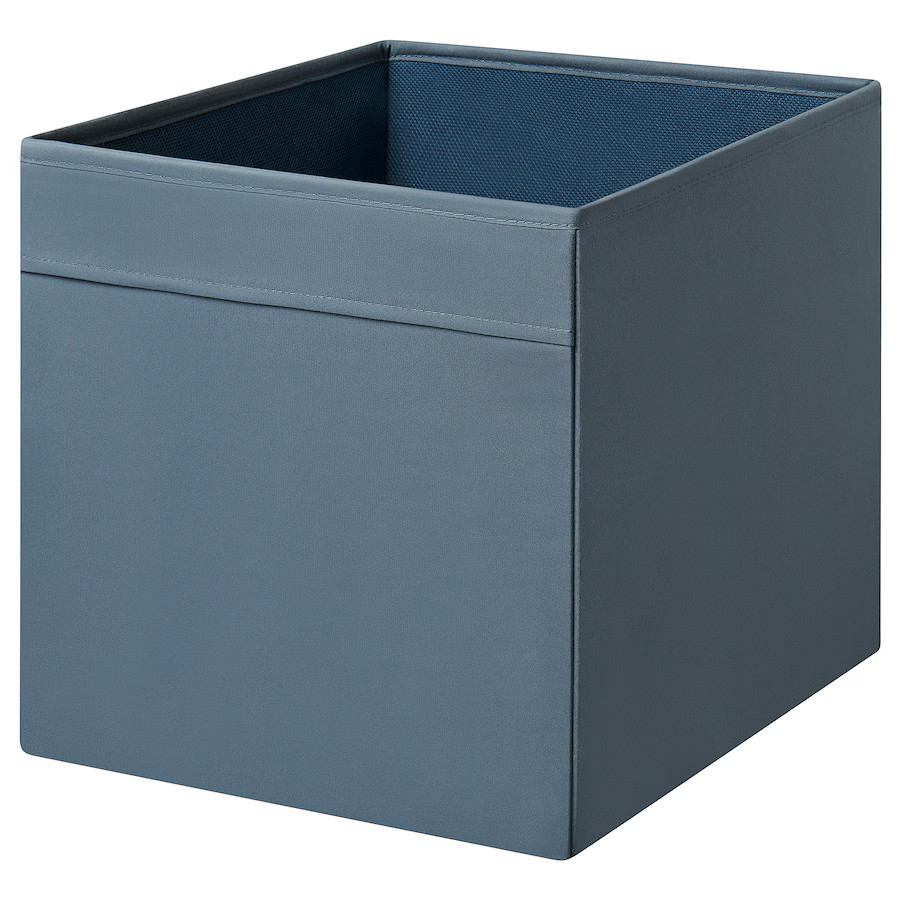}
    \end{minipage}
    \begin{minipage}{0.075\textwidth}
        \centering        \includegraphics[width=1.2cm,height=1.2cm,keepaspectratio]{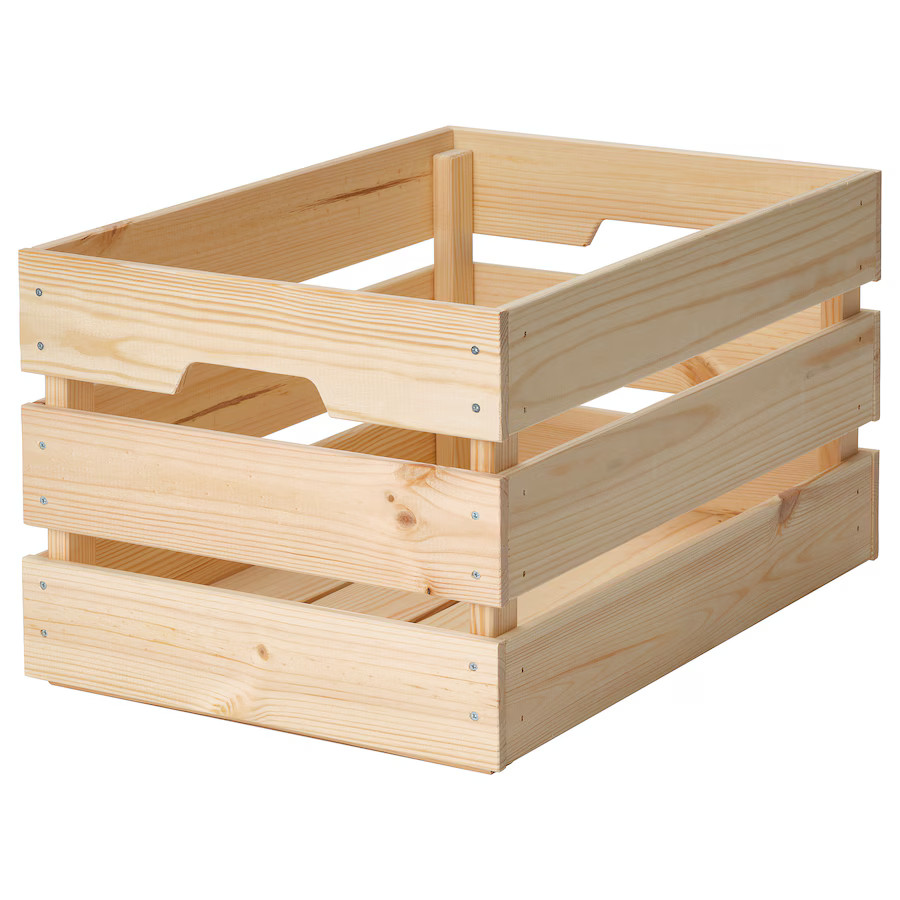}
    \end{minipage}
    
    \begin{minipage}{0.075\textwidth}
        \centering        \includegraphics[width=1cm,height=1cm,keepaspectratio]{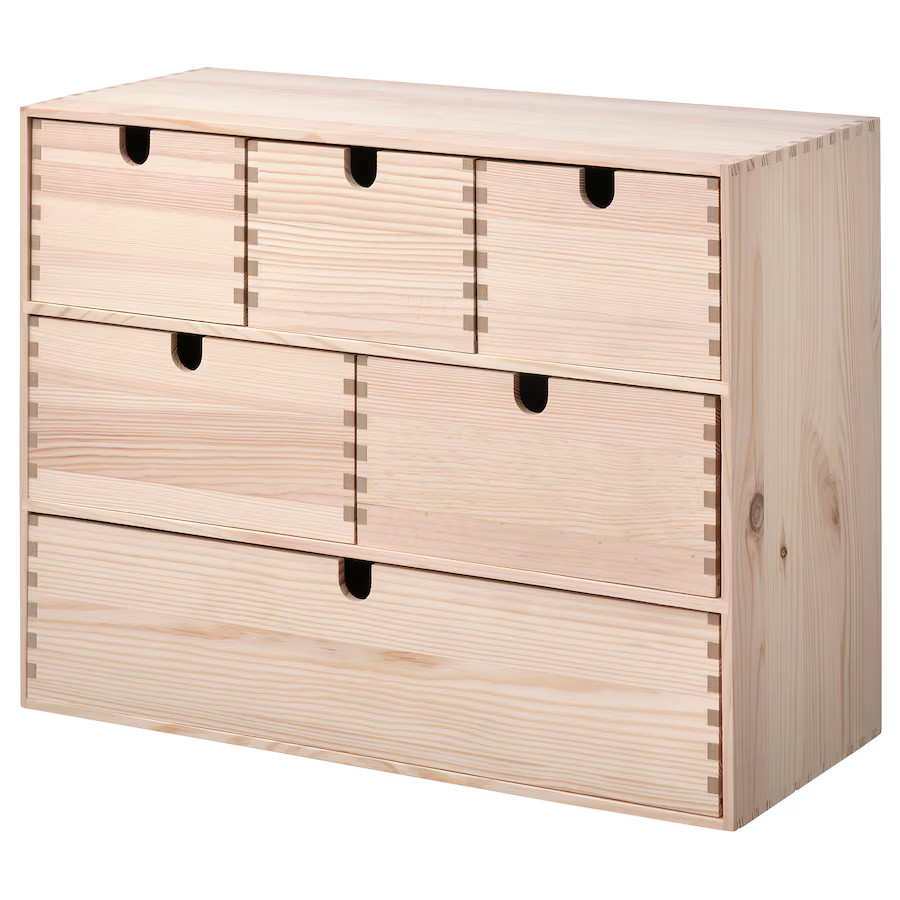}
    \end{minipage}
    \begin{minipage}{0.065\textwidth}
        \centering
        \includegraphics[width=\textwidth,height=0.9cm,keepaspectratio]{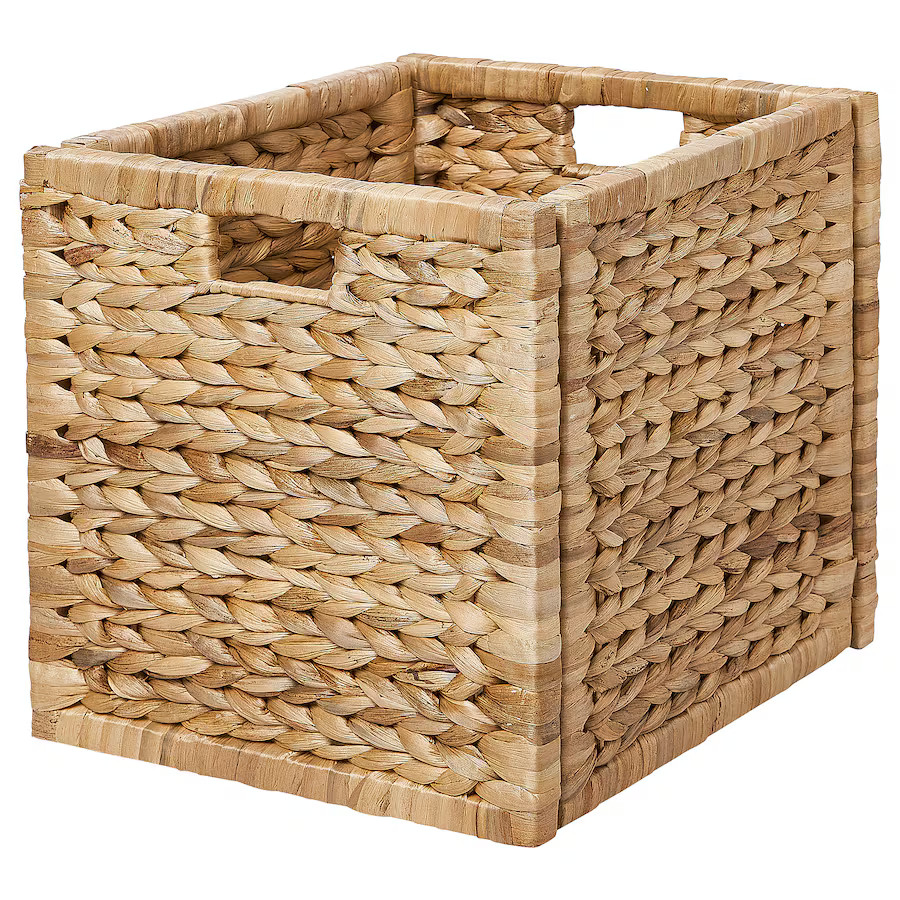}
    \end{minipage}
    \begin{minipage}{0.13\textwidth}
        \centering
    \includegraphics[width=1.15cm,height=1.45cm]{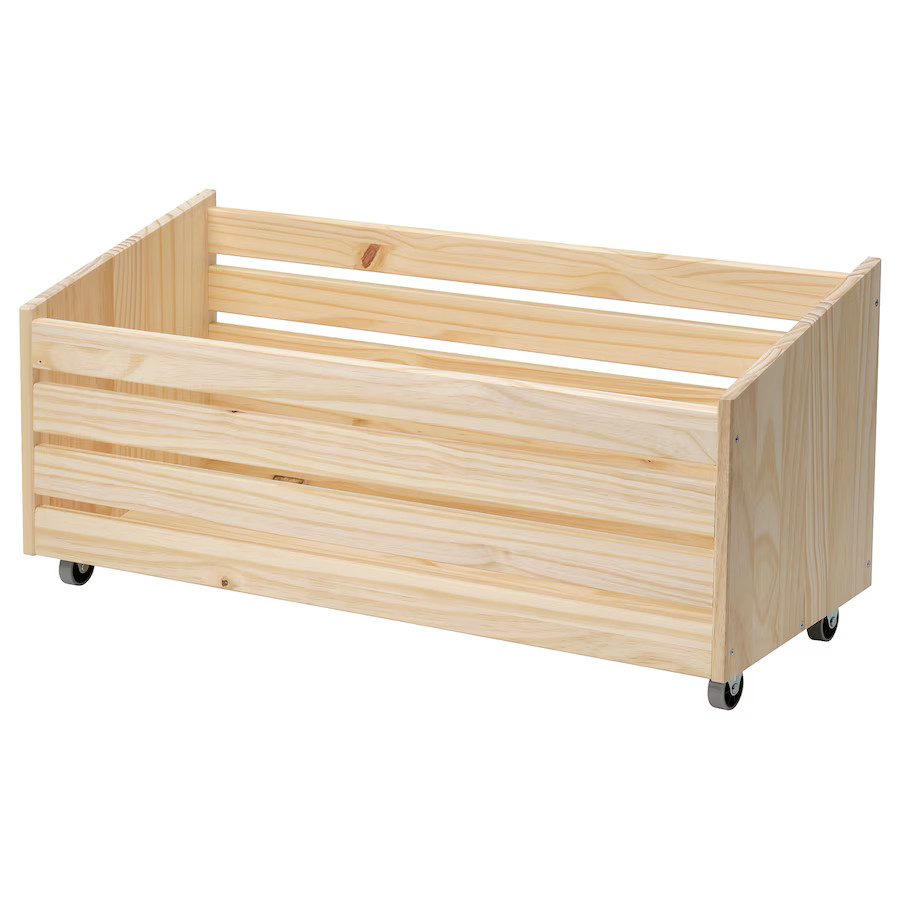}
    \end{minipage}
    \begin{minipage}{0.13\textwidth}
        \centering
        \includegraphics[width=1.9cm,height=1.25cm]{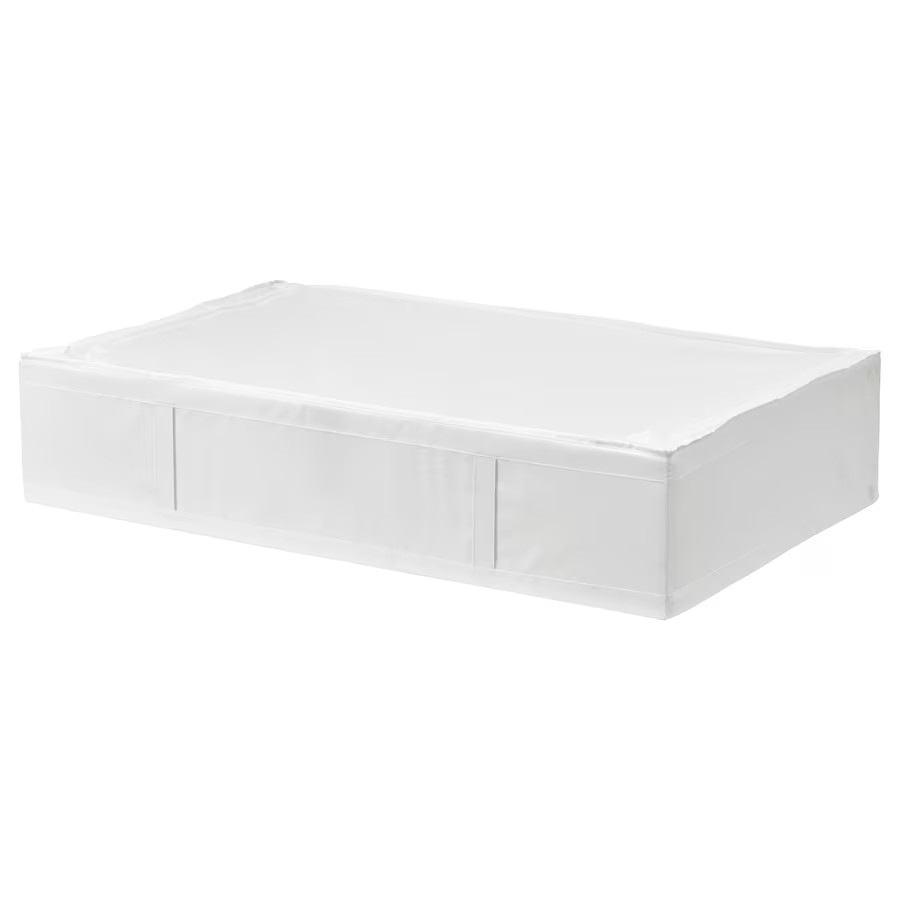}
    \end{minipage}
    \begin{minipage}{0.055\textwidth}
        \centering
        \includegraphics[width=\textwidth,height=1cm,keepaspectratio]{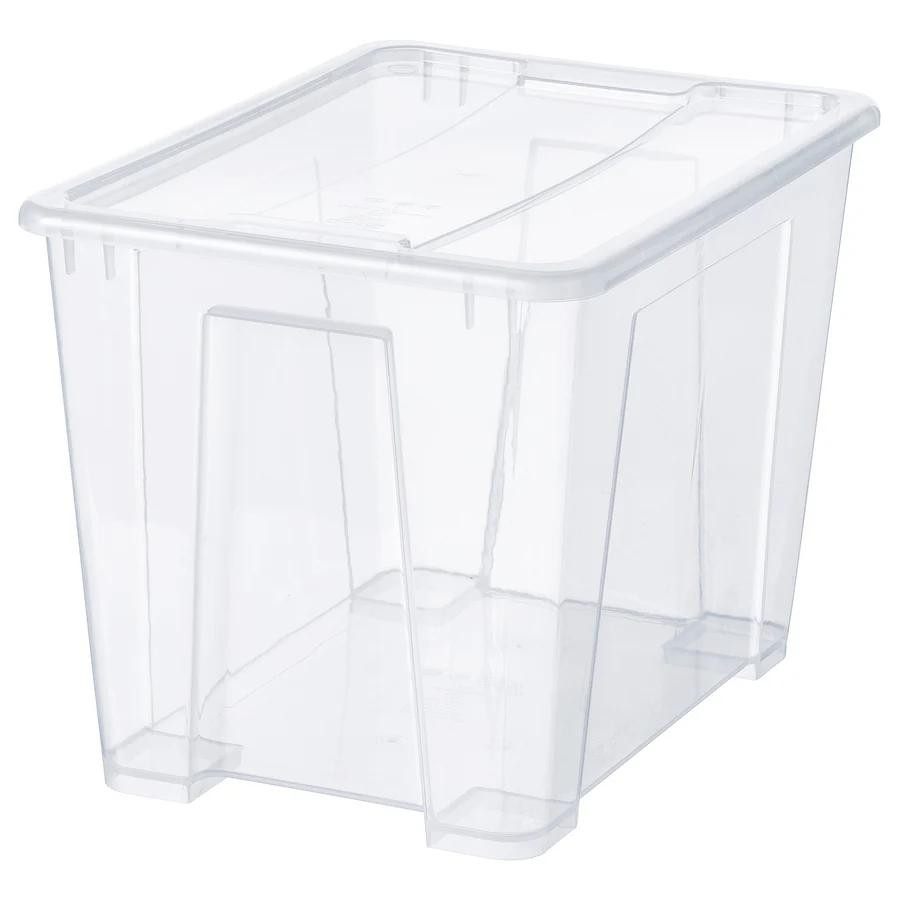}
    \end{minipage}
    \caption{The objects used in our MR6D dataset. These objects are the same as those used in the MTevent dataset \cite{mtevent} which features stereo-event and monocular RGB camera recordings with 6D pose annotations and 3D bounding boxes for all moving objects. MTevent primary goal is to support 3D bounding box detection of high-speed moving objects using event data, which does not overlap with the purpose of MR6D.}
    
    \label{fig:mr6d_objects}
\end{figure}

%% file: figures/fig_manual_anno.tex
\begin{figure}
    \centering
    \includegraphics[height=4cm, trim = 0 350 0 350, clip]{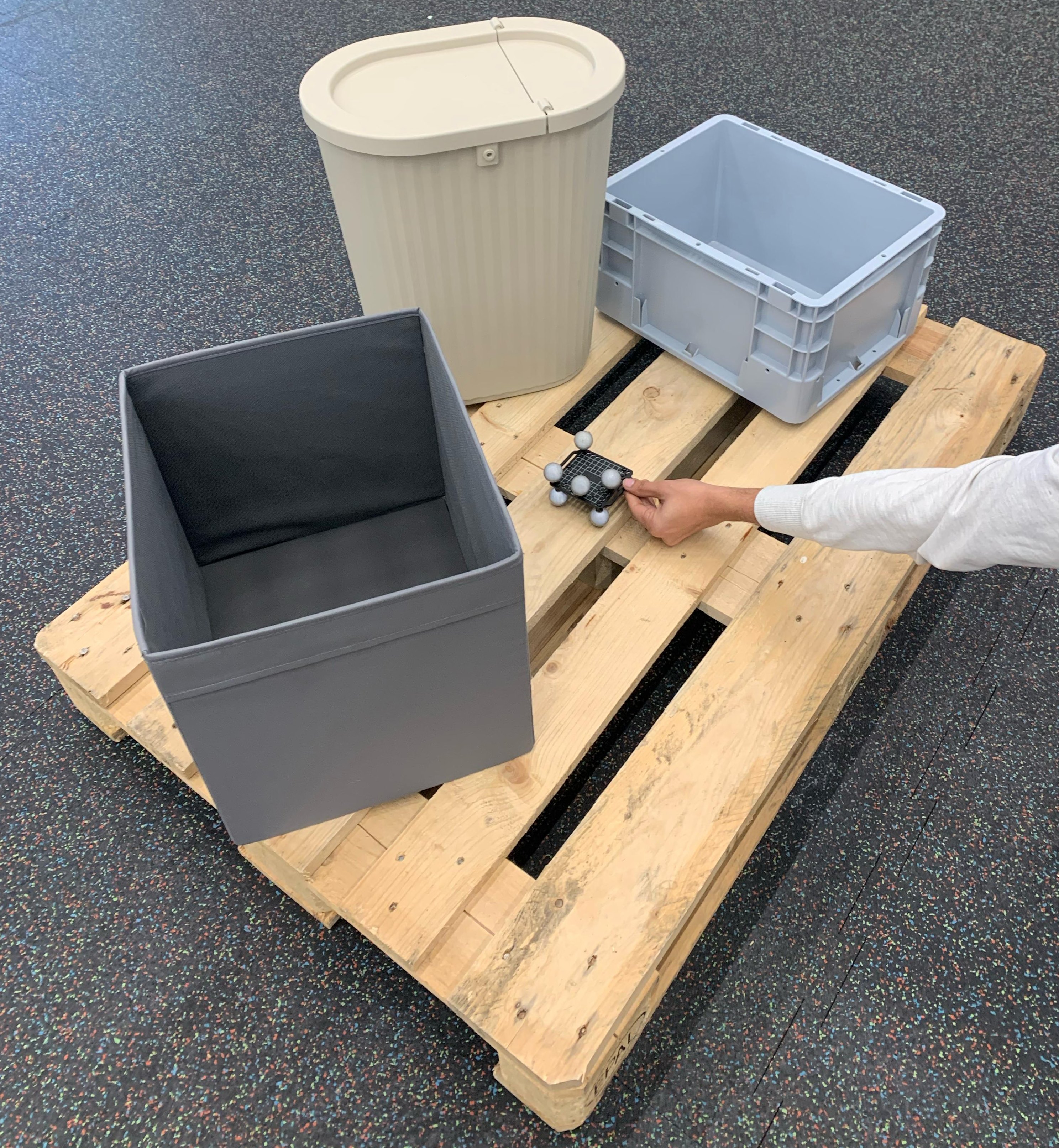}
    \includegraphics[height=4cm]{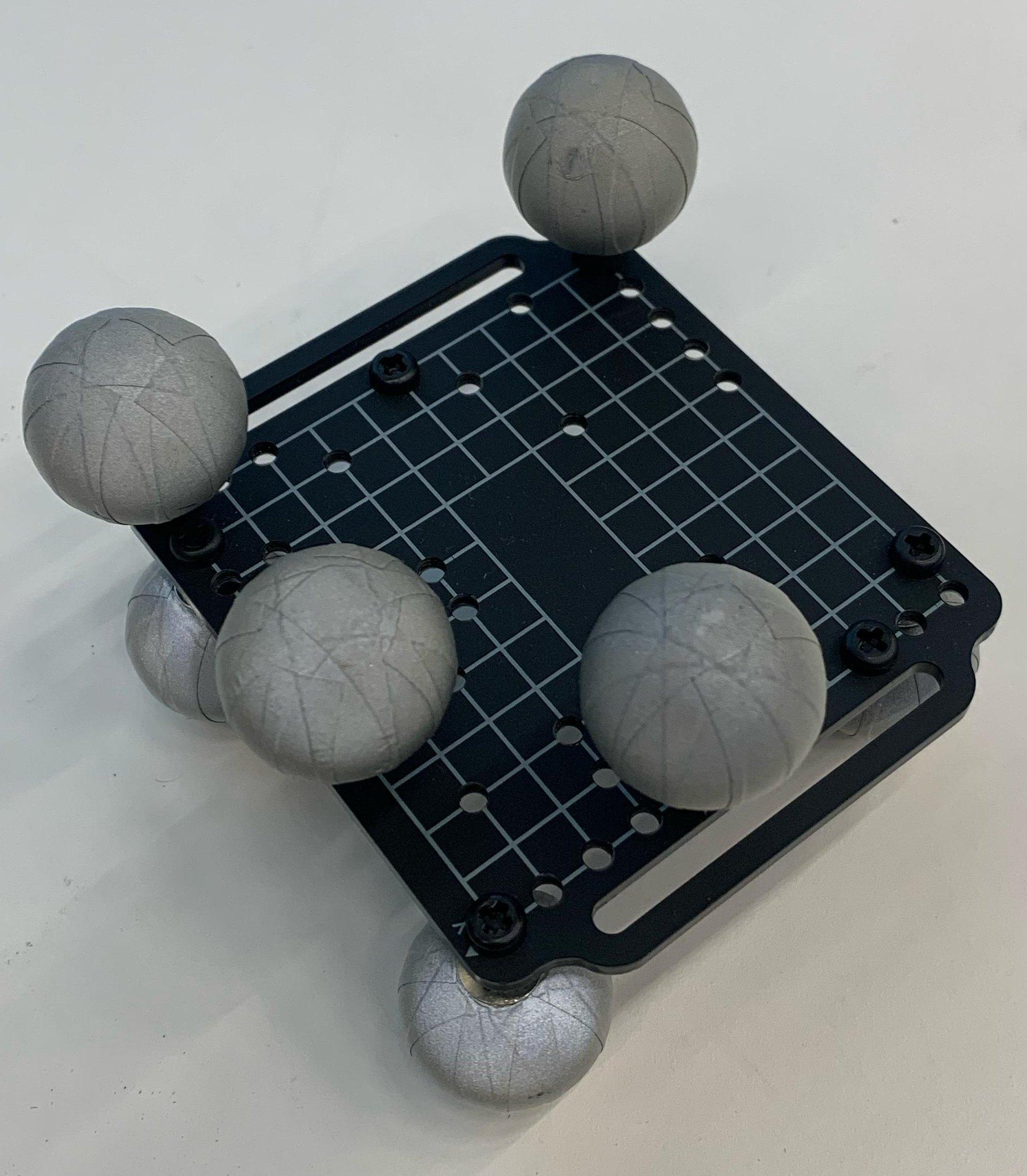}
    \caption{Annotating object 6D pose in the validation subset. After placing all objects in the scene, their initial 6D poses are recorded in the world frame using a marking object positioned at the highest Z-axis point and aligned with the object’s orientation. Since this initial capture is not perfectly accurate, the poses are later manually refined using the BOP manual annotation tool. The refinement is performed for the entire scene at once, requiring minimal effort—even complex scenes can be adjusted in under a minute.}
    \label{fig:manual_anno}
\end{figure}

%% file: figures/fig_collection_process.tex
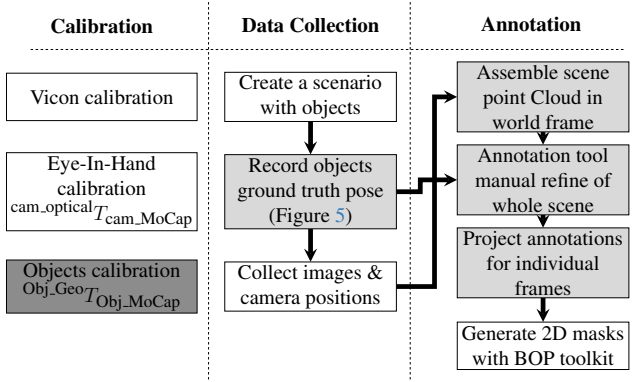
\begin{figure}
    \centering

\resizebox{\linewidth}{!}{
\begin{tikzpicture}[node distance=2cm]
\draw[dashed] (-3,2) -- (22,2);

\draw[dashed] (4.5,4) -- (4.5,-12);

\draw[dashed] (13,4) -- (13,-12);

    \node (box1) [startstop] {\parbox{7.8cm}{\centering \Huge Vicon calibration}};
    \node (box2) [startstop, below of=box1, yshift=-2cm] {
        \parbox{7.8cm}{\centering \Huge Eye-In-Hand calibration\\
        $^{\text{cam\_optical}}T_{\text{cam\_MoCap}}$ }};
    \node (box3) [startstop, below of=box2, yshift=-2cm , fill=gray!90] {
        \parbox{7.8cm}{\centering \Huge Objects  calibration\\
        $^{\text{Obj\_Geo}} T _{\text{Obj\_MoCap}} $ }};

    \node at ([yshift=2cm]box1.north) {\Huge \textbf{Calibration}};
    \node (box3) [startstop, right of=box1, xshift=6.8cm] {\parbox{7cm}{\centering \Huge Create a scenario with objects}};
    \node (box4) [startstop, below of=box3, yshift=-2cm , fill=gray!30] { \parbox{7cm}{\centering \Huge Record objects ground truth pose (Figure \ref{fig:manual_anno})}};
    \node (box5) [startstop, below of=box4, yshift=-2cm] { \parbox{7cm}{\centering \Huge Collect images \& camera positions}};
    \node at ([yshift=2cm]box3.north){\Huge \textbf{Data Collection}}; 

    \draw [arrow] (box3) -- (box4);
    \draw [arrow] (box4) -- (box5);
    
    \node (box6)  [startstop, right of=box3, xshift=7.8cm, fill=gray!30] { \parbox{7cm}{\centering \Huge Assemble scene point Cloud in world frame}};
    \node (box8)  [startstop, below of=box6, yshift=-1.5cm, fill=gray!30]  { \parbox{7cm}{\centering \Huge Annotation tool manual refine of whole scene}};
    \node (box9)  [startstop, below of=box8, yshift=-1.5cm , fill=gray!30]  { \parbox{7cm}{\centering \Huge Project annotations for individual frames}};
    \node (box10) [startstop, below of=box9, yshift=-1.5cm]  { \parbox{7cm}{\centering \Huge Generate 2D masks with BOP toolkit}};
      
    \node at ([yshift=1.6cm,xshift=-0.5cm]box6.north){\Huge \textbf{Annotation}}; 
    \draw [arrow] (box6) -- (box8);
    \draw [arrow] (box8) -- (box9);
    \draw [arrow] (box9) -- (box10);
    \draw [arrow] (box5.east) -- ++(1.6cm,0) |- (box6.west);

    \draw [arrow] (box4.east) -- ++(1.0cm,0) |- (box8.west);

\end{tikzpicture}
}

    \caption{Dataset collection process for the validation and dynamic subsets. Light gray indicates steps specific to the validation subset, while dark gray denotes steps exclusive to the dynamic subset.}
    \label{fig:collection_process}
\end{figure}

%% file: figures/fig_data_statistics.tex
\begin{figure*}
    \centering
    \includegraphics[height=3.7cm]{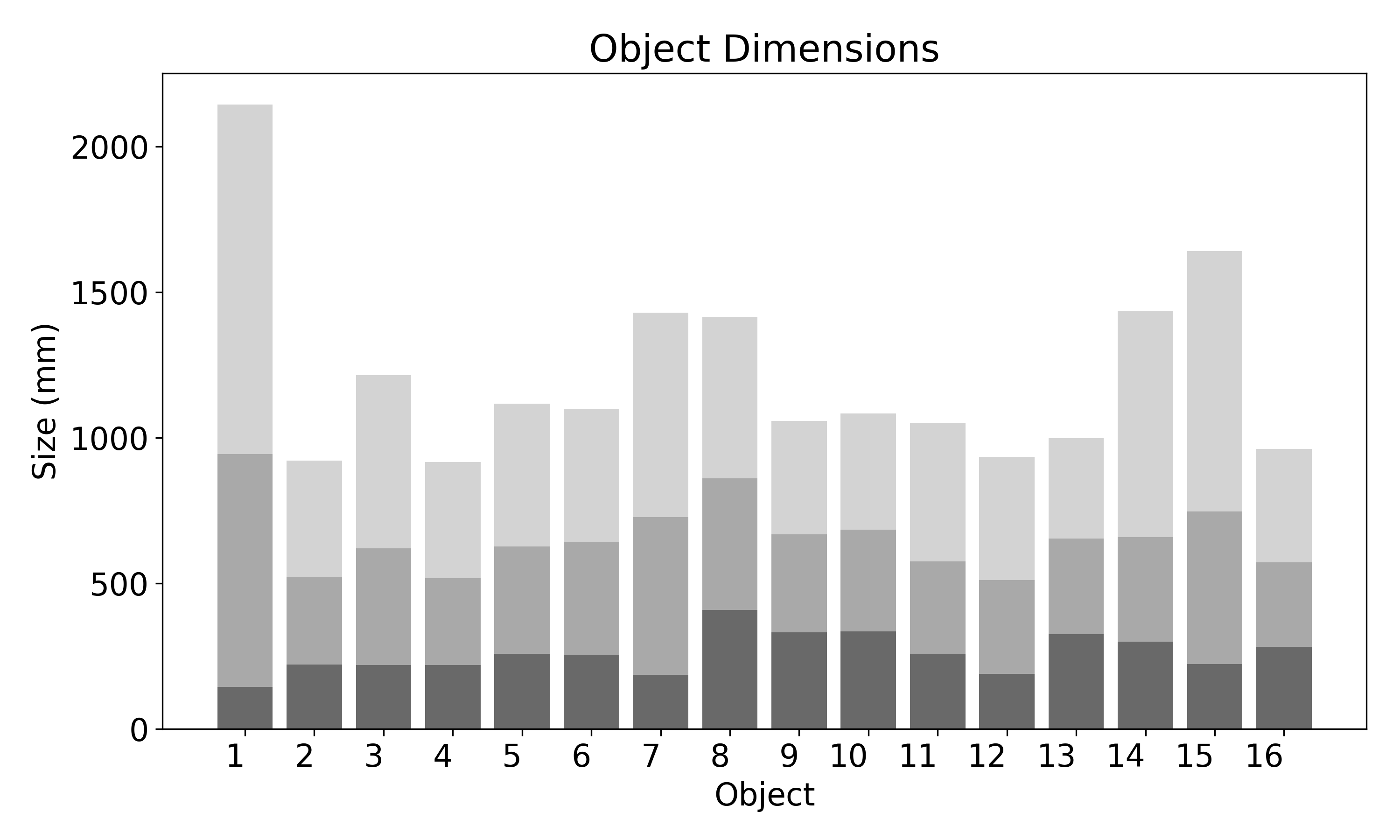}
    \includegraphics[height=3.7cm]{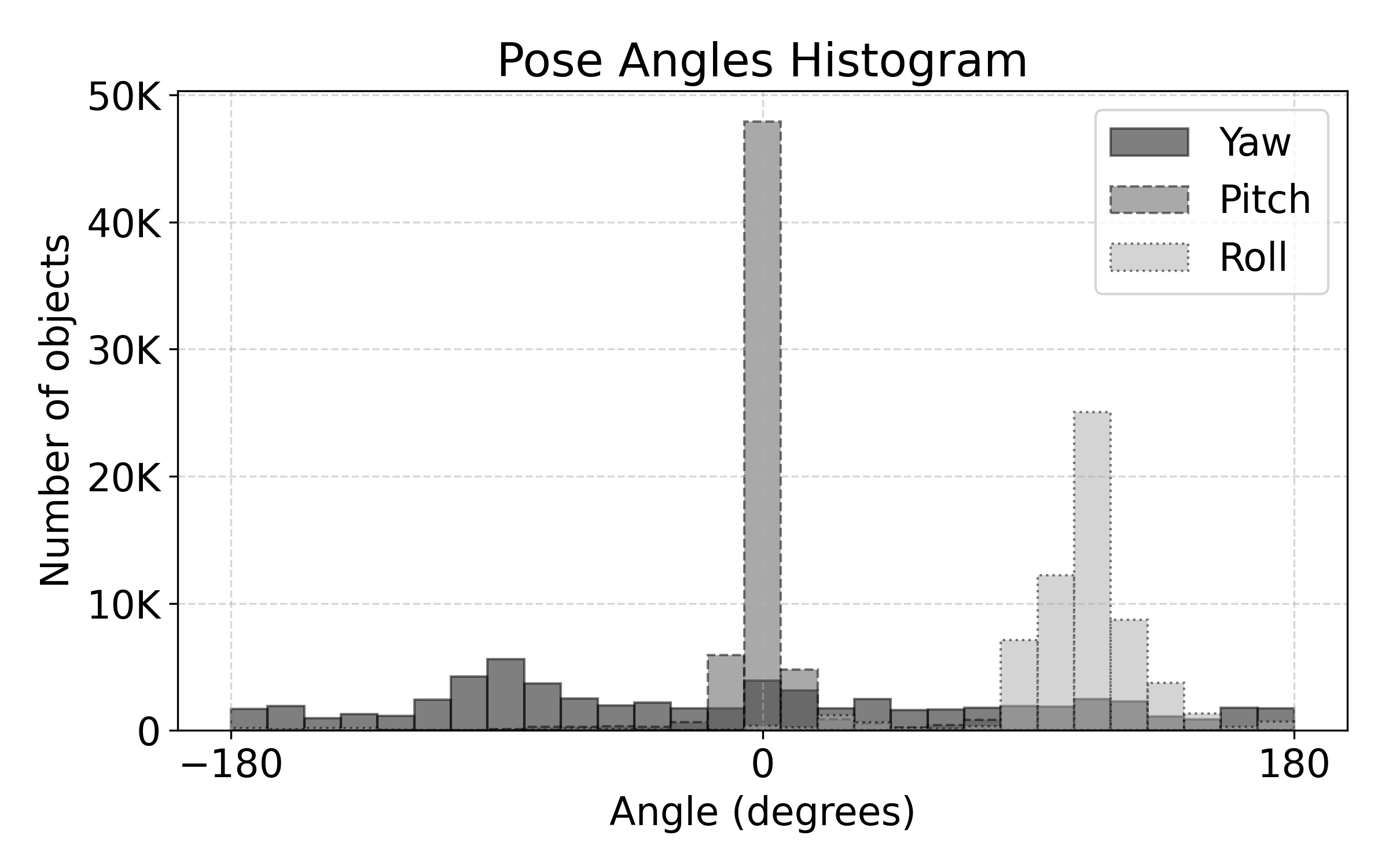} \\
    \includegraphics[height=3.7cm]{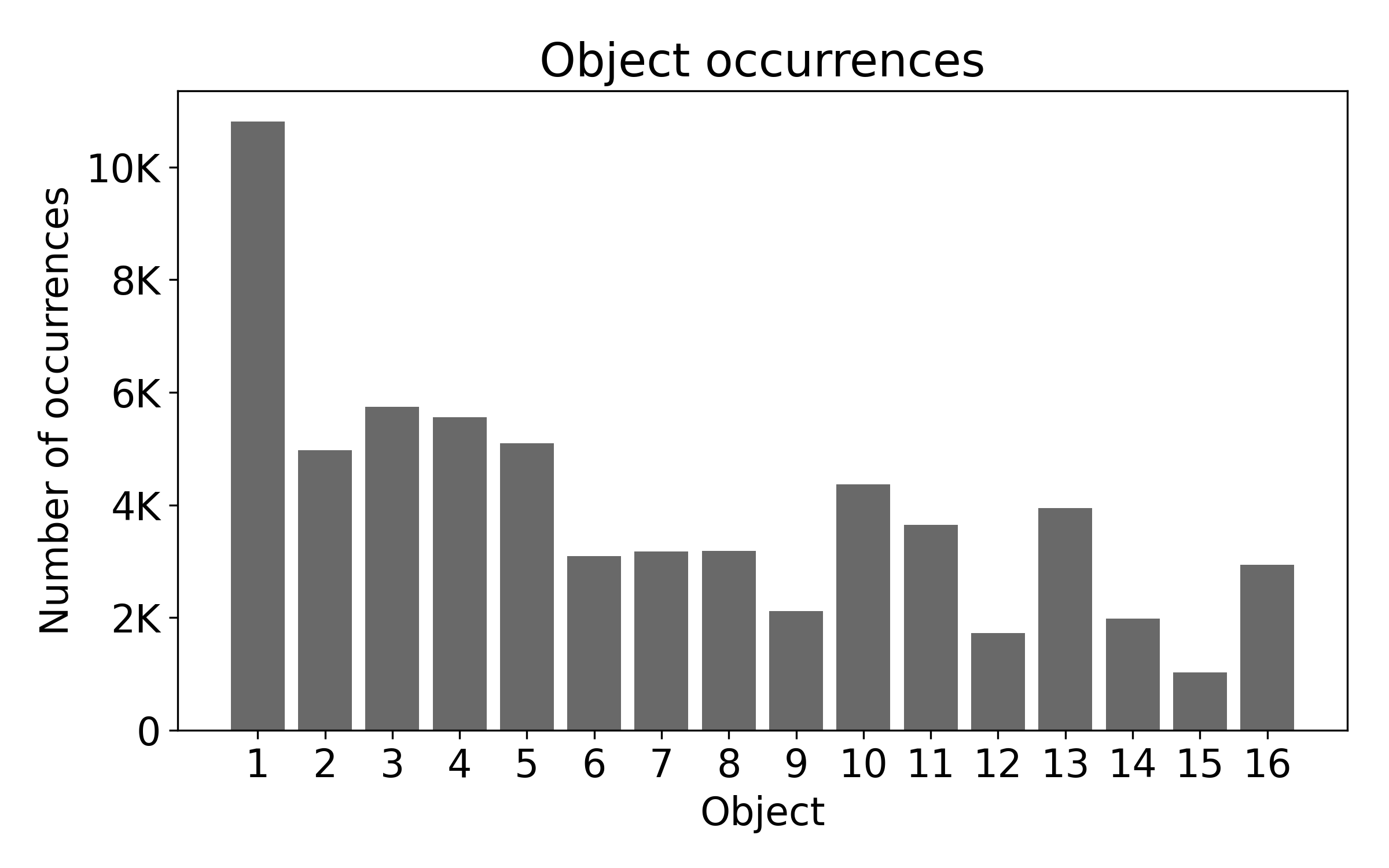}
    \includegraphics[height=3.7cm]{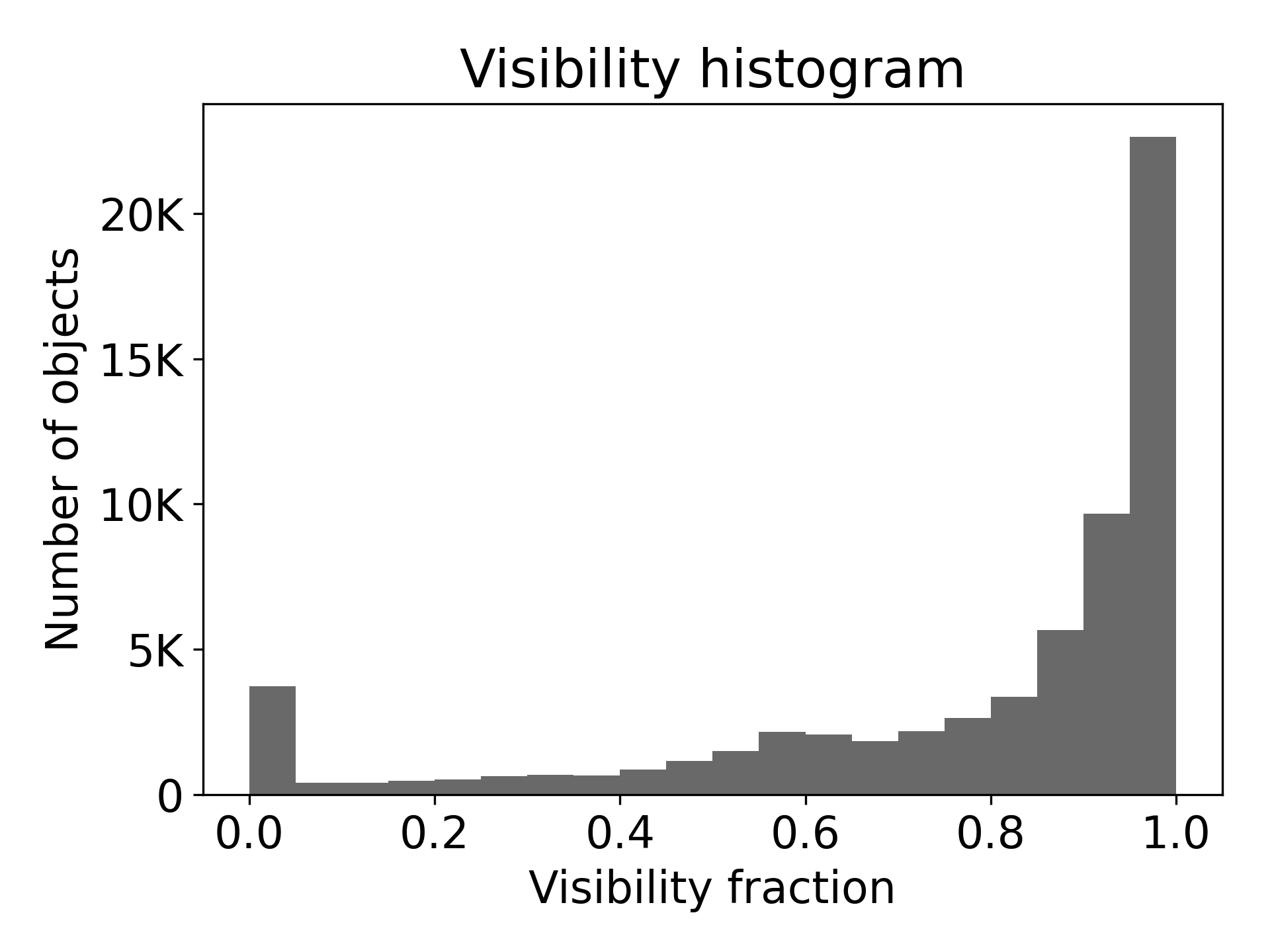}
    \includegraphics[height=3.7cm]{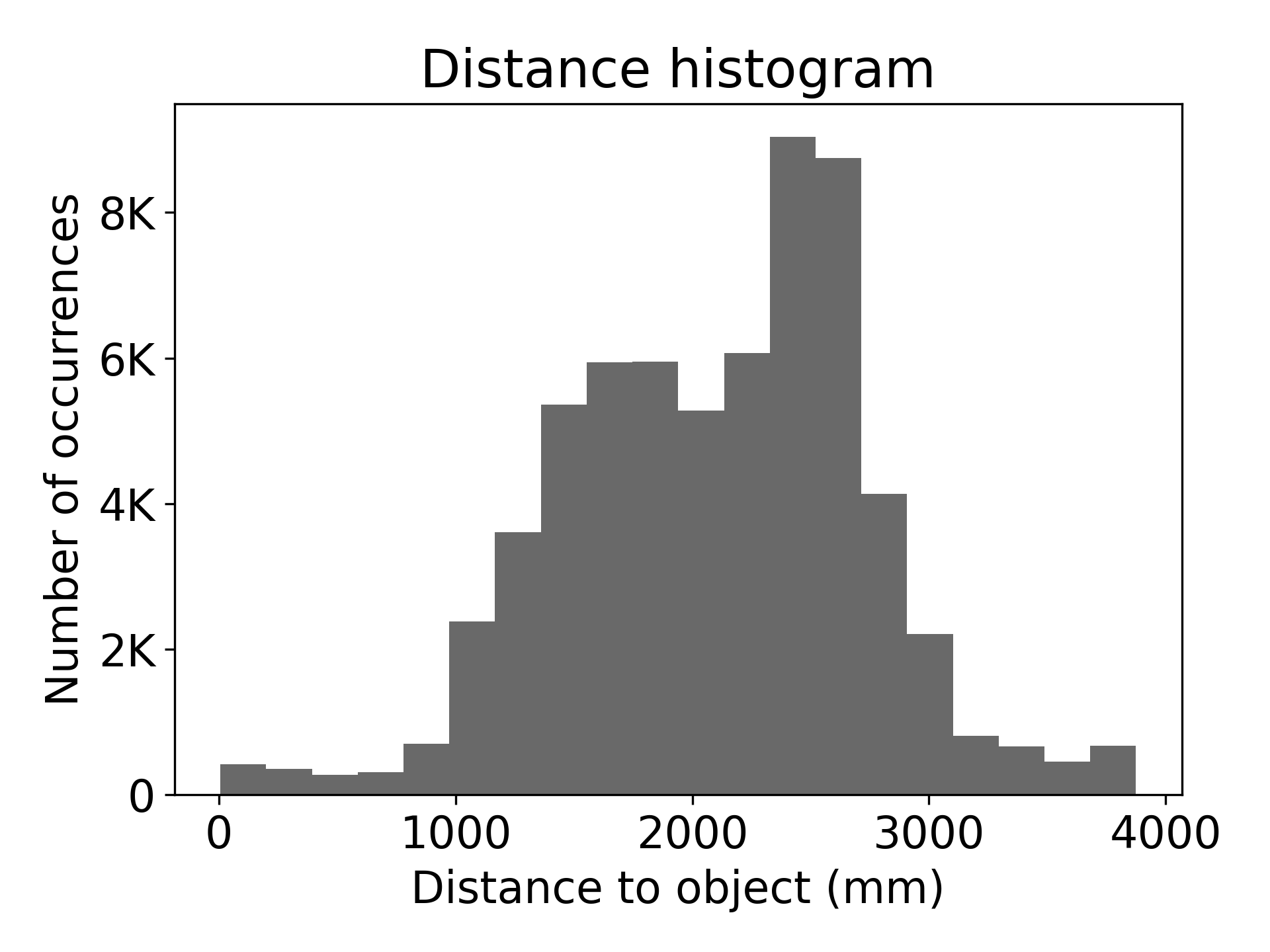}
    \caption{Data statistics. The first plot shows object dimensions (length, breadth, height), highlighting the minimum, median, and maximum values for each object. The second plot displays a pose angle histogram, representing the distribution of object orientations. The third plot shows the number of occurrences per object, with Object 1 (Euro Pallet) appearing more frequently due to its frequent inclusion in the O³dyn test subset. The fourth plot presents a visibility histogram, where zero visibility indicates the object is outside the camera frame but still has 6D annotations. The fifth plot shows a histogram of object distances from the camera.}
    \label{fig:data_statistics}
\end{figure*}

%% file: figures/fig_dataset_subsets.tex
\begin{figure*}
    \centering
    \includegraphics[width=0.3\textwidth]{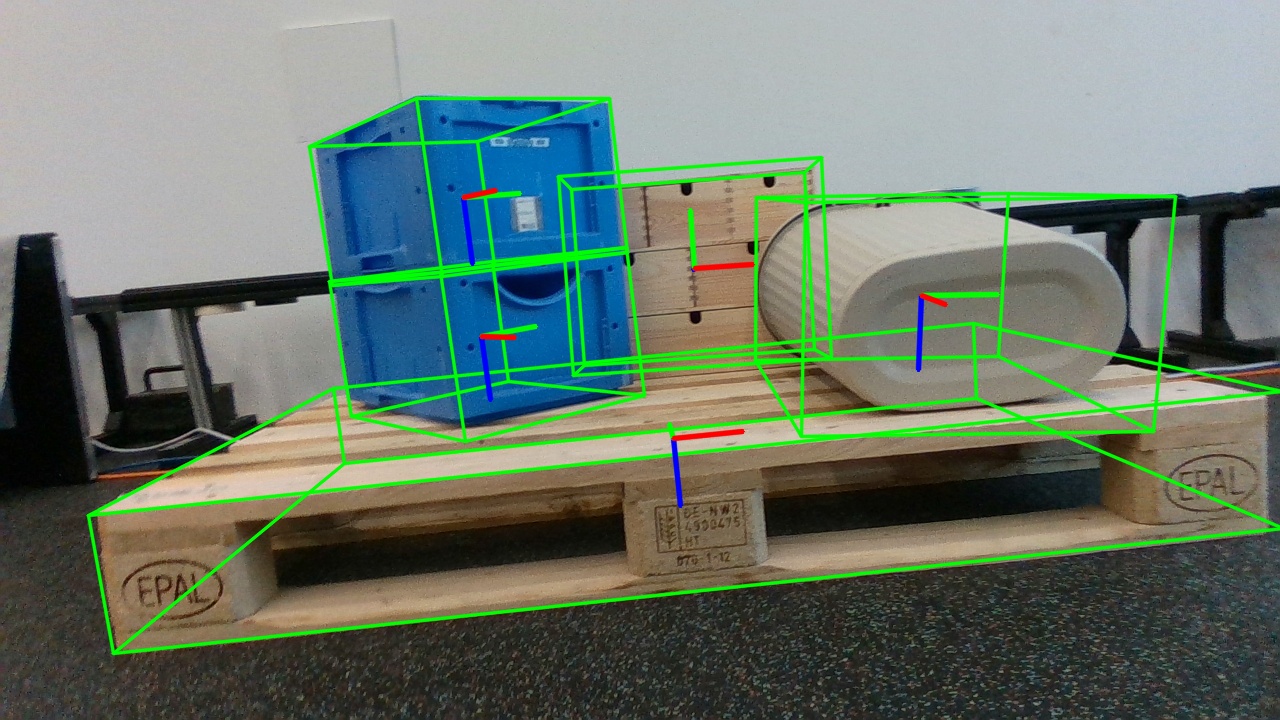}
    \includegraphics[width=0.3\textwidth]{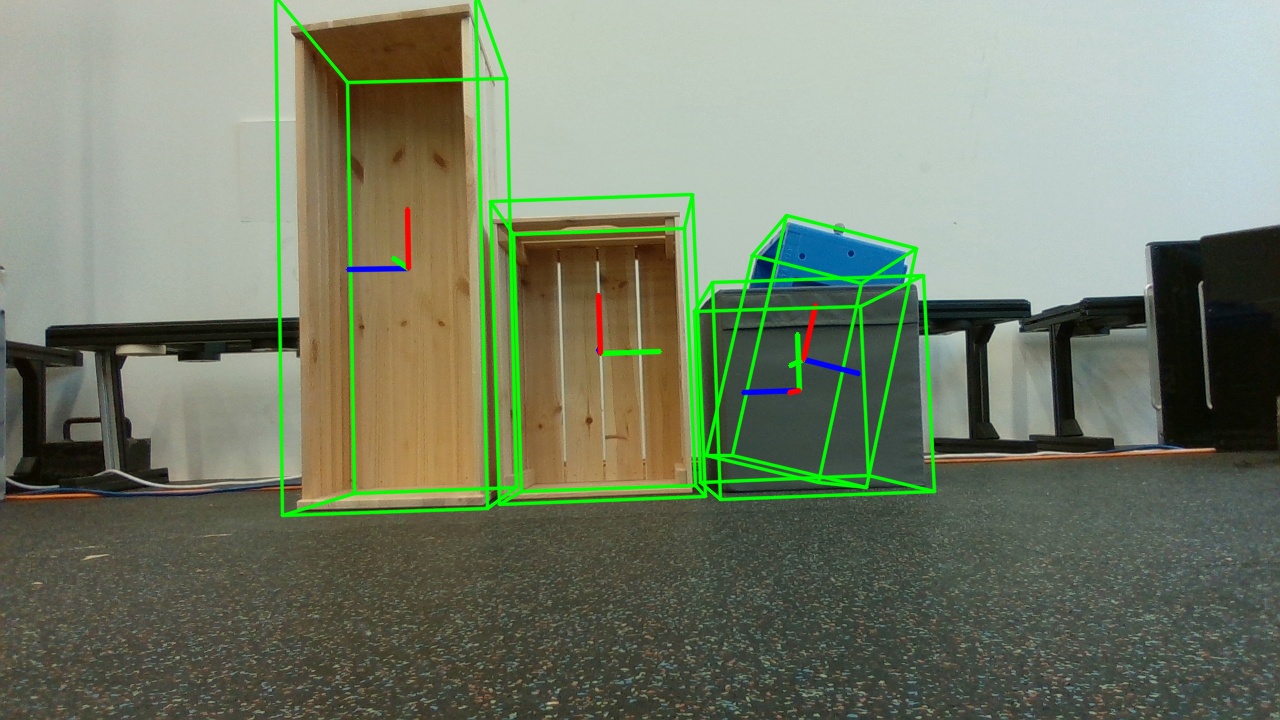}
    \includegraphics[width=0.3\textwidth]{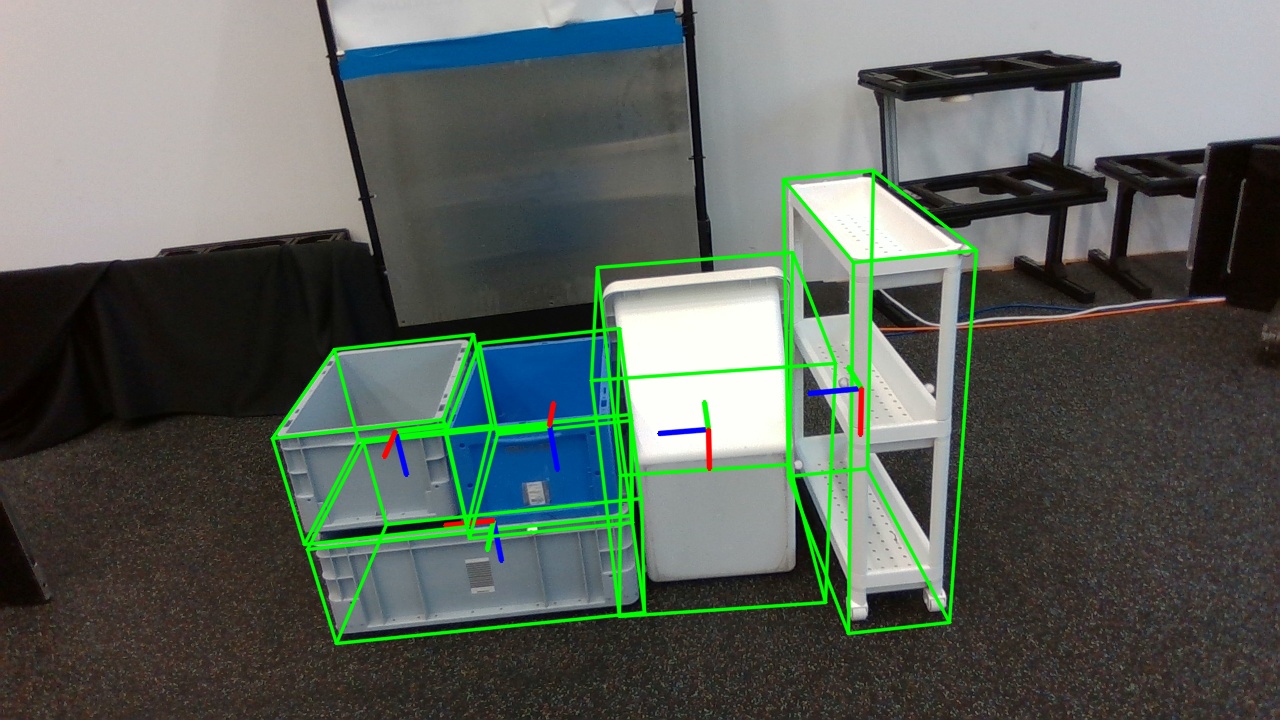} \\
    \vspace{0.1cm}
    \centering
    \includegraphics[width=0.3\textwidth]{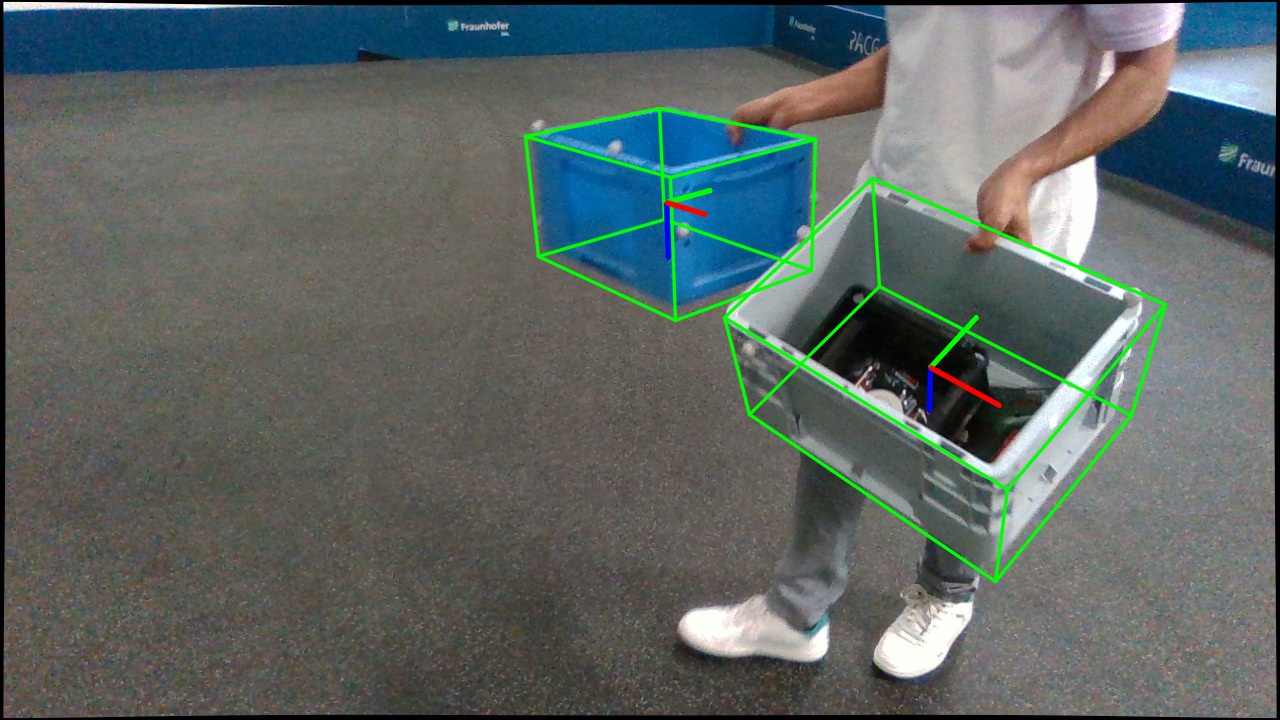}
    \includegraphics[width=0.3\textwidth]{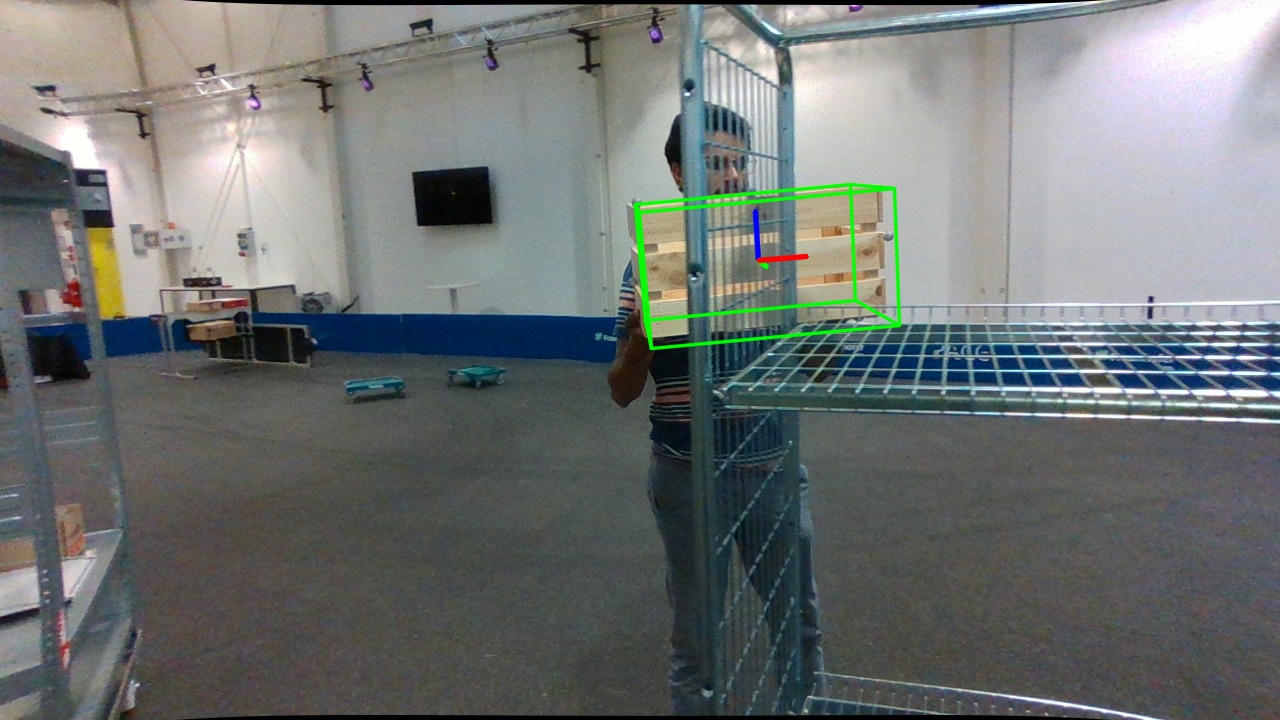}
    \includegraphics[width=0.3\textwidth]{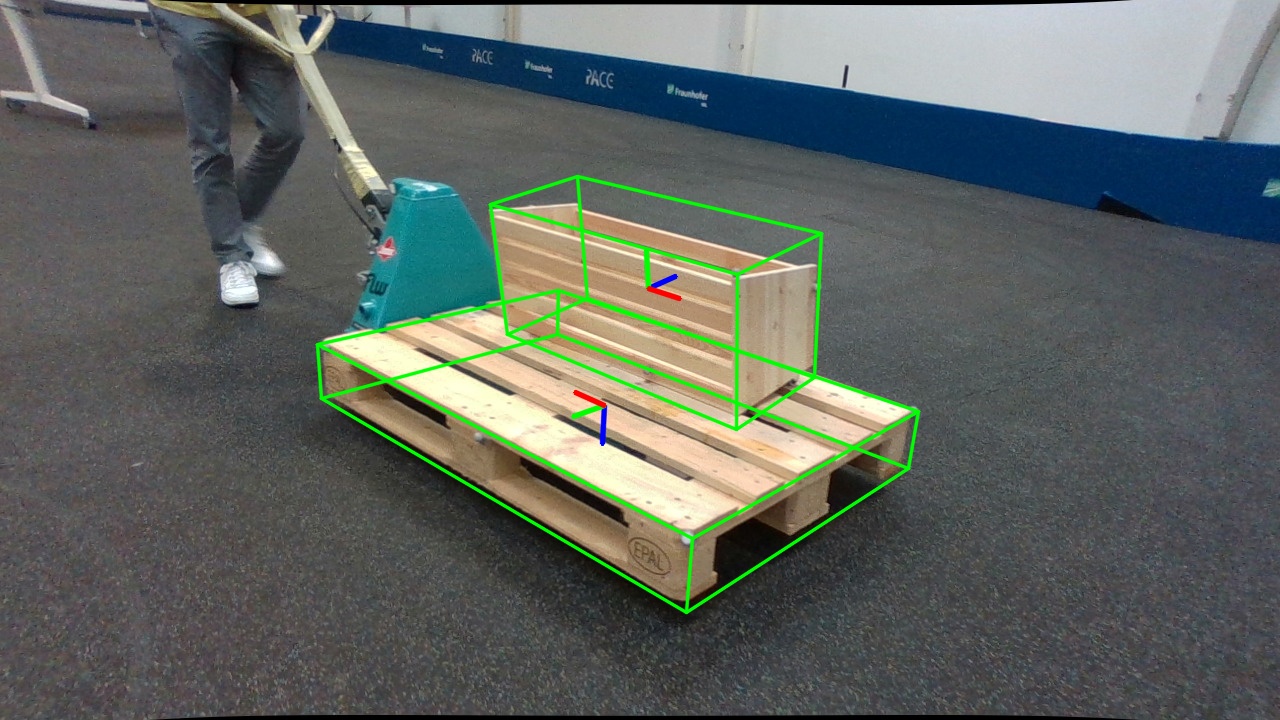} \\
    \vspace{0.1cm}
    \centering
    \includegraphics[width=0.3\textwidth]{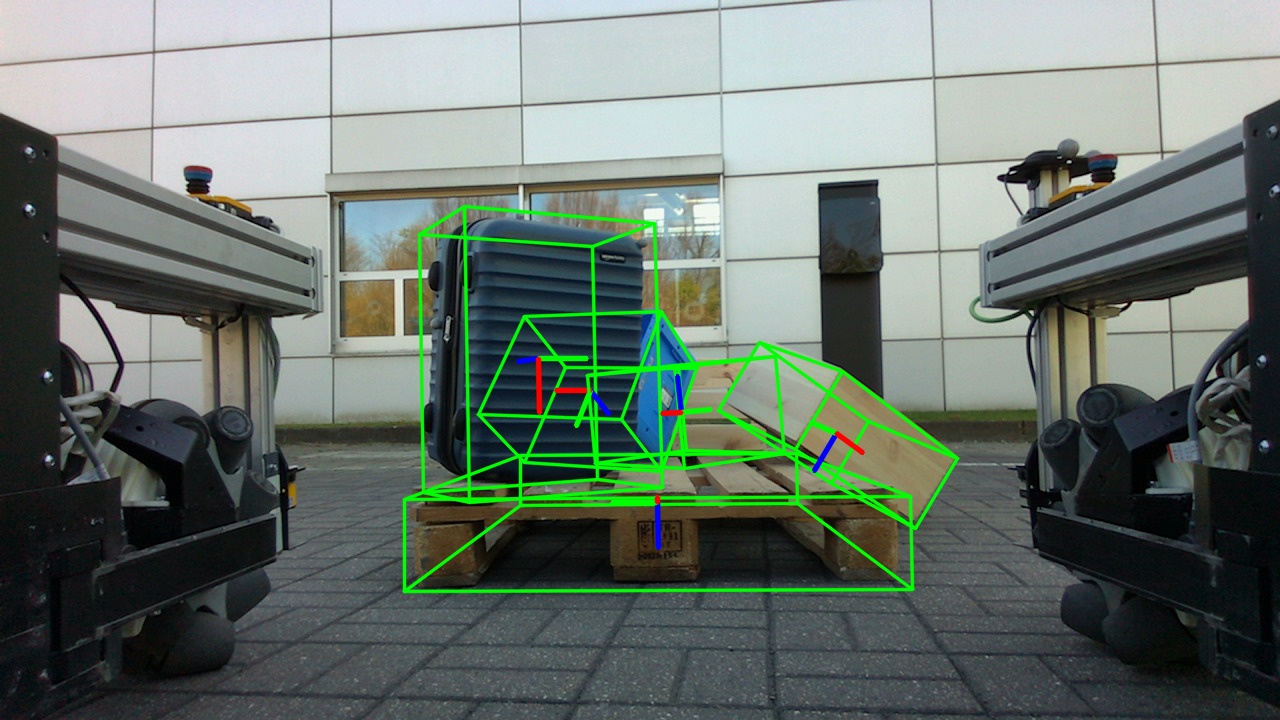}
    \includegraphics[width=0.3\textwidth]{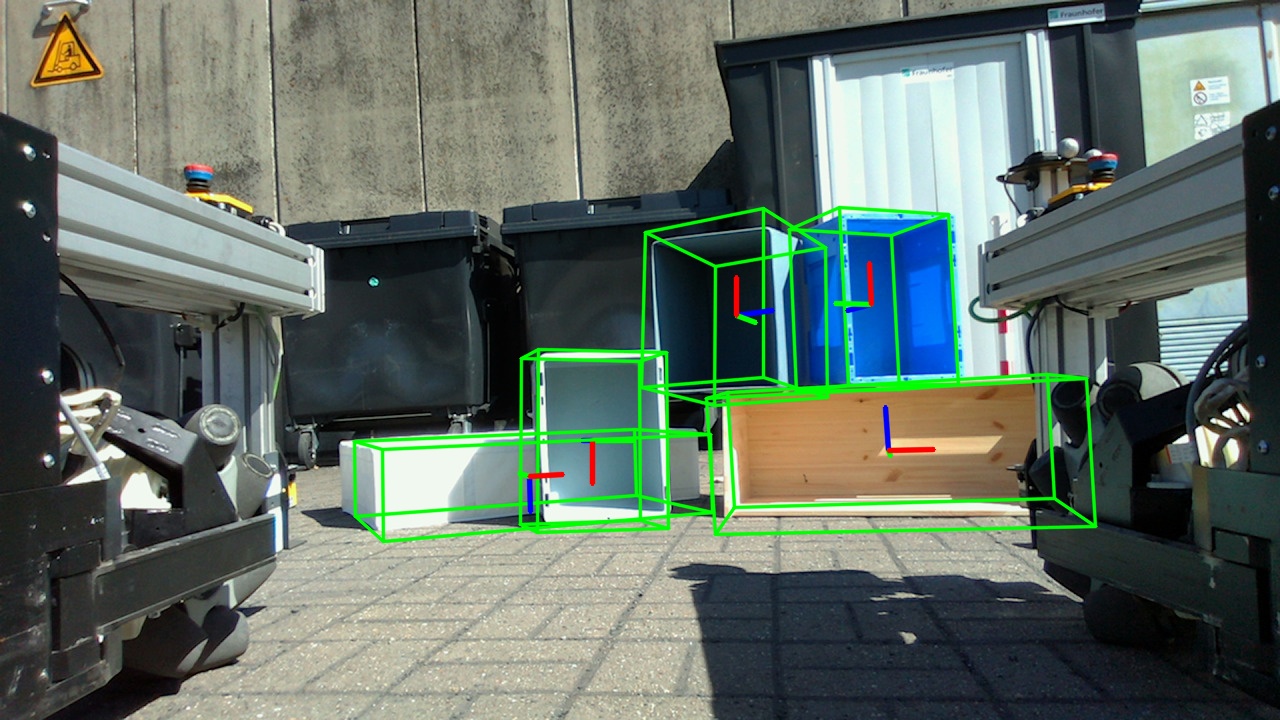}
    \includegraphics[width=0.3\textwidth]{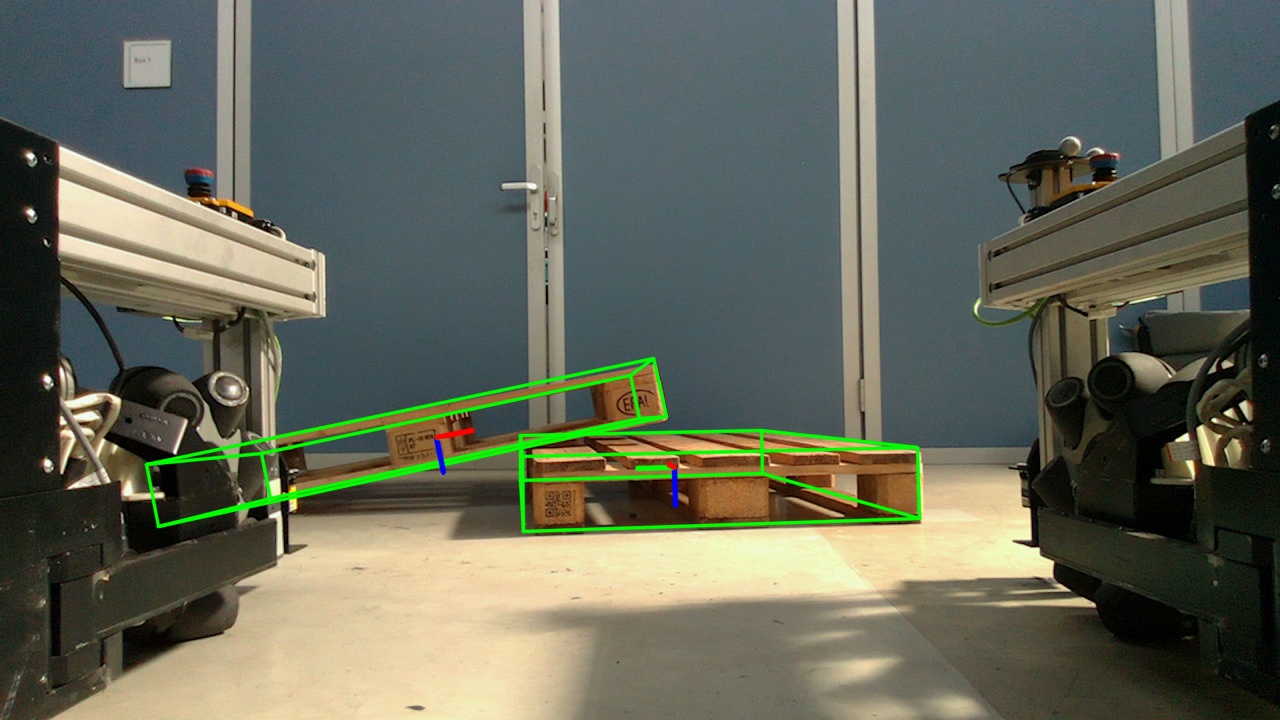} \\
    \vspace{0.1cm}
    \centering
    \includegraphics[width=0.3\textwidth]{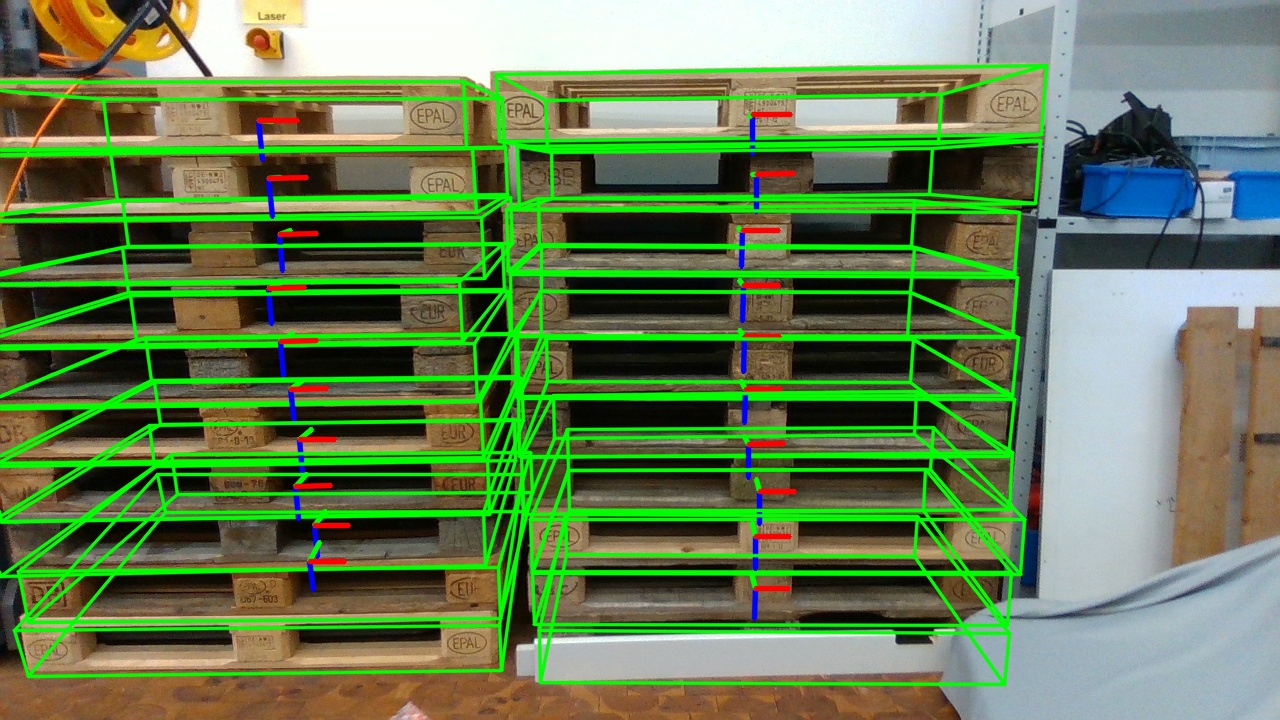}
    \includegraphics[width=0.3\textwidth]{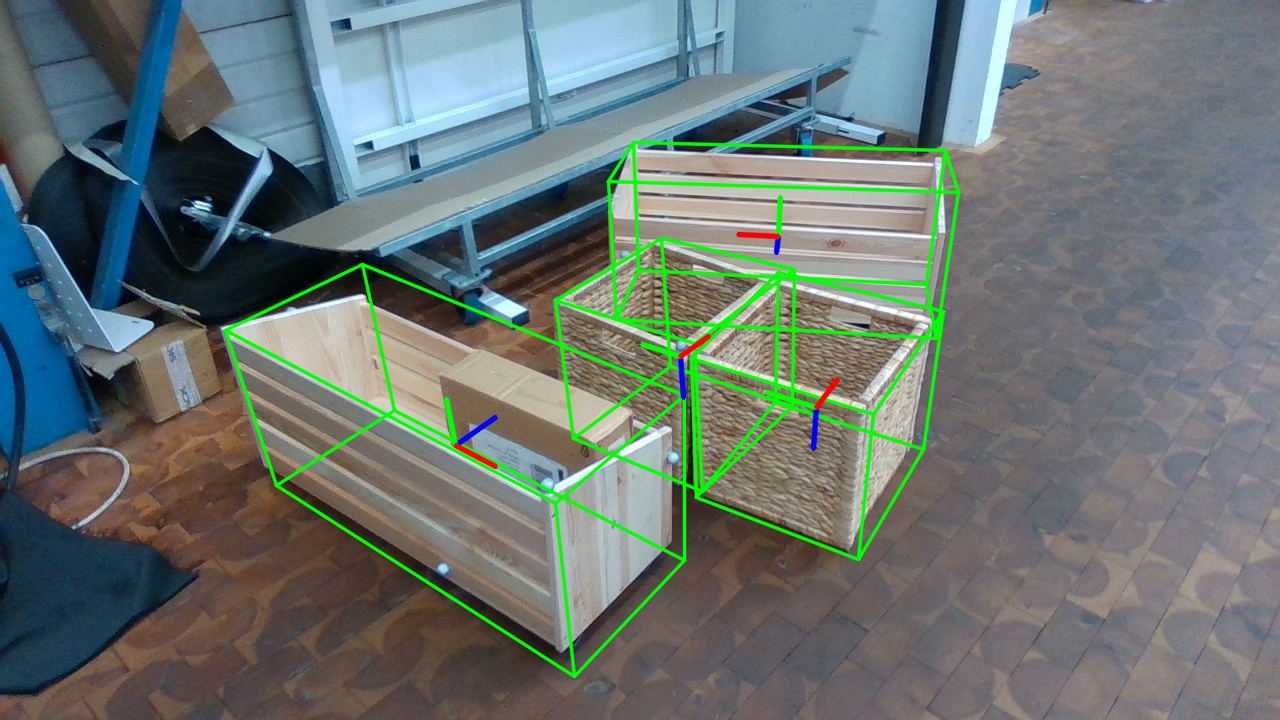}
    \includegraphics[width=0.3\textwidth]{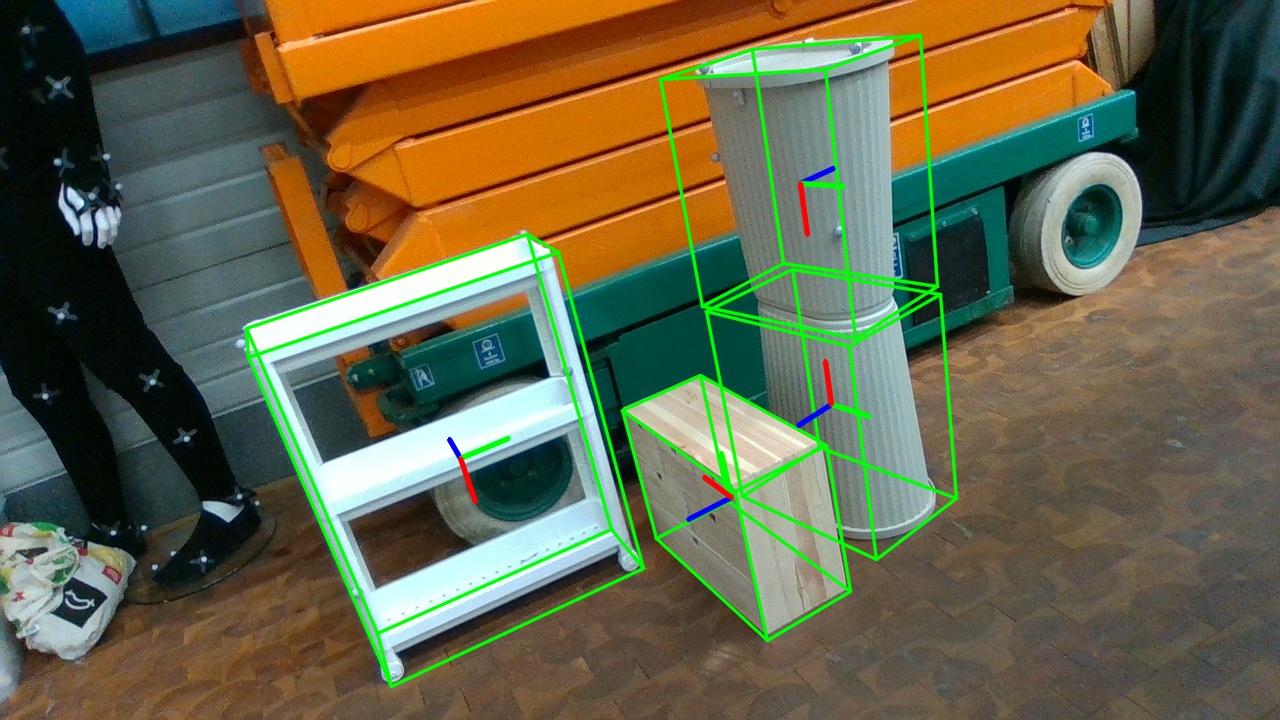}
    \caption{MR6D comprises four subsets, each shown in a row: static validation scenes (Row 1), human-manipulated dynamic scenes (Row 2), O³dyn scenes from low–mounted AGV camera perspectives (Row 3), and mobile robot-like scenes with varied viewpoints and occlusions (Row 4). Bounding boxes can be larger than the visible parts of the objects, covering elements such as wheels or curved surfaces.}
    \label{fig:dataset_subsets}
\end{figure*}

%% file: sec/4_evaluation.tex
\section{Evaluation}
\label{sec:evaluation}

\input{figures/fig_eval_table}
\input{figures/fig_foundationpose_vis}
\input{figures/fig_failure_cases}

In this section, we evaluate our dataset using pipelines for unseen object 6D pose estimation and present a qualitative analysis highlighting challenging cases.

\subsection{Evaluation with Pipelines of Unseen Objects}

We conduct our evaluation using pipelines designed for 6D pose estimation of unseen objects.
The first evaluation uses FoundationPose~\cite{foundationpose}, a \emph{model-based} method, together with ground-truth masks. 
The second evaluation employs a fully unseen pipeline that uses Centroid Triplet Loss (CTL)~\cite{ctl}, a \emph{model-free} method, to generate 2D segmentation masks of unseen objects. 
CTL is chosen over DINOv2~\cite{dinov2} due to its superior segmentation performance. 
For this pipeline, 4--6 reference images of each object are captured and used as query images.
We evaluate performance using BOP metrics~\cite{bop2020}.

Table~\ref{tab:eval_bop} presents the results for each subset individually. The GT-masks+FoundationPose pipeline achieves an Average Recall (AR) of 0.3462 on average across the 3 test subsets, whereas CTL-generated masks reduce AR to 0.1841. These results highlight that while 6D pose models show some generalization to mobile robotics applications, performance is still influenced by 2D segmentation quality. Figure~\ref{fig:eval_vis} visualizes the ground-truth and predicted projections for one frame from each subset.

\subsection{Analysis of Challenging Cases}

Figure~\ref{fig:failure_cases} presents examples where 6D pose predictions struggle, even when using ground-truth masks. One primary challenge is misidentification due to occlusion, as shown in the left image, where the model predicts the pose of a nearby object instead of the intended target. This issue is less frequent when both objects are viewed from an angle that provides sufficient visibility. In the middle image, two similar objects are closely stacked, which can lead to incorrect pose estimation for one or both objects. The right image shows another challenge: when a dominant face is visible alongside a nearby object with a similar texture, the model may predict a completely incorrect orientation.

%% file: figures/fig_eval_table.tex
\begin{table*}[htbp]
    \centering
    \caption{6D pose evaluation results on the different subsets of MR6D}
    \label{tab:eval_bop}
    \begin{tabular}{lcccccccc}
        \toprule
        Subset & \multicolumn{4}{c}{GT-Masks + FoundationPose} & \multicolumn{4}{c}{CTL-DoUnseen-Masks + FoundationPose} \\
        \cmidrule(lr){2-5} \cmidrule(lr){6-9}
                & AR & AR$_{\text{MSPD}}$ & AR$_{\text{MSSD}}$ & AR$_{\text{VSD}}$ 
                & AR & AR$_{\text{MSPD}}$ & AR$_{\text{MSSD}}$ & AR$_{\text{VSD}}$ \\
        \midrule
        Validation   & 0.4432 & 0.4422 & 0.3794 & 0.5081 & 0.2275 & 0.2207 & 0.1977 & 0.2641 \\
        \midrule
        Test MR      & 0.3722 & 0.3736 & 0.3399 & 0.4030 & 0.2005 & 0.2052 & 0.1791 & 0.2173 \\
        Test O³dyn   & 0.2927 & 0.2936 & 0.3082 & 0.2763 & 0.1673 & 0.1652 & 0.1742 & 0.1625 \\
        Test Dynamic & 0.3738 & 0.3821 & 0.3149 & 0.4244 & 0.1843 & 0.1804 & 0.1632 & 0.2094 \\
        Test Avg.    & 0.3462 & 0.3498 & 0.3210 & 0.3679 & 0.1841 & 0.1836 & 0.1722 & 0.1964 \\
        \bottomrule
    \end{tabular}
\end{table*}

%% file: figures/fig_foundationpose_vis.tex
\begin{figure*}
    \centering
    \includegraphics[height=2.4cm]{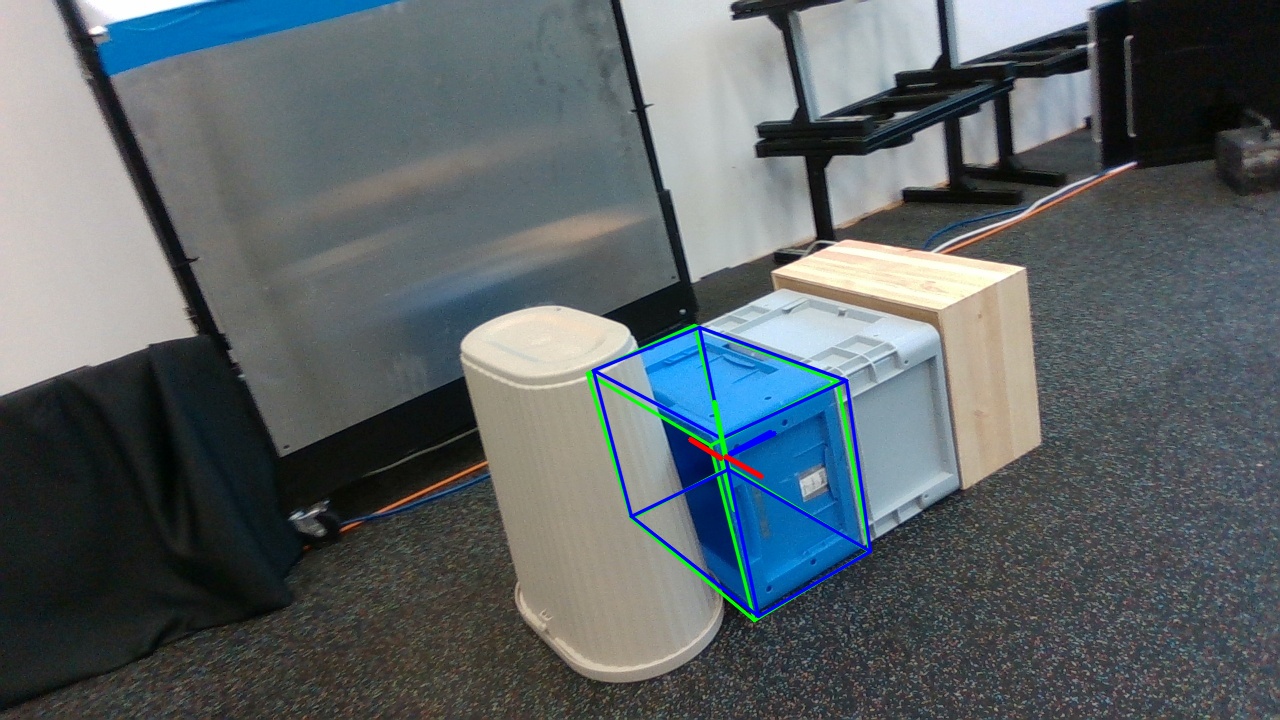}
    \includegraphics[height=2.4cm]{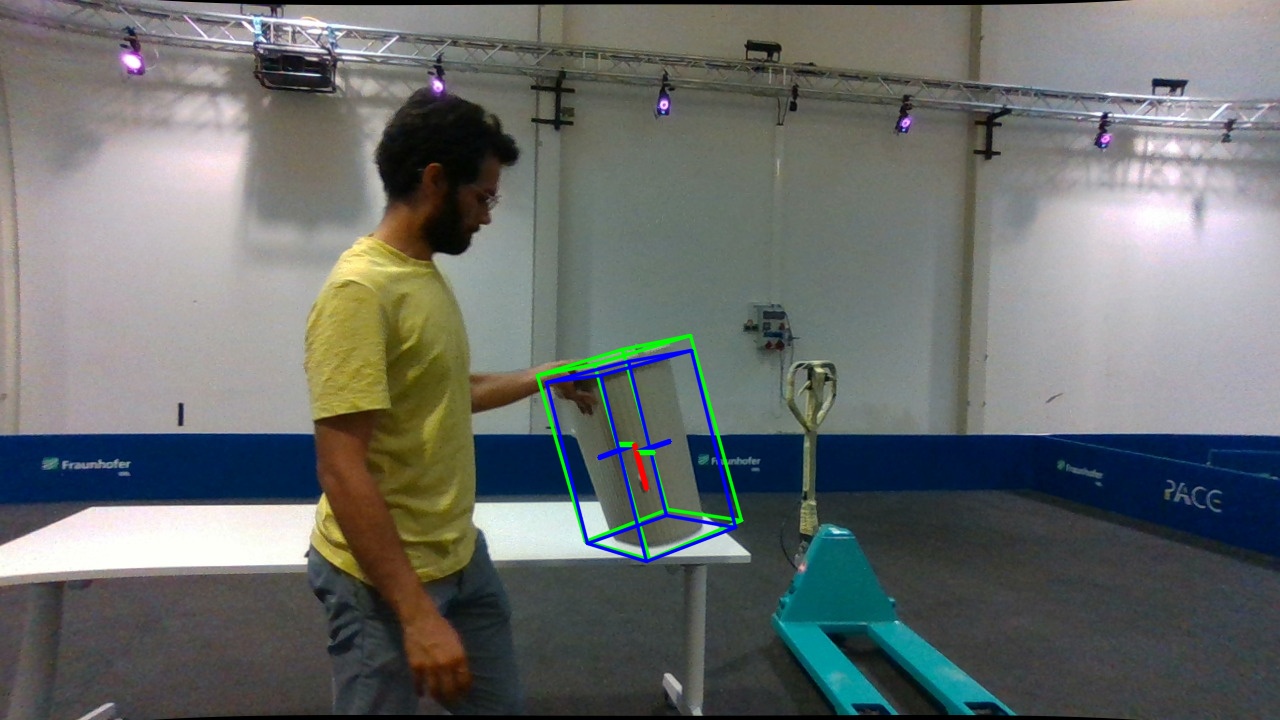}
    \includegraphics[height=2.4cm]{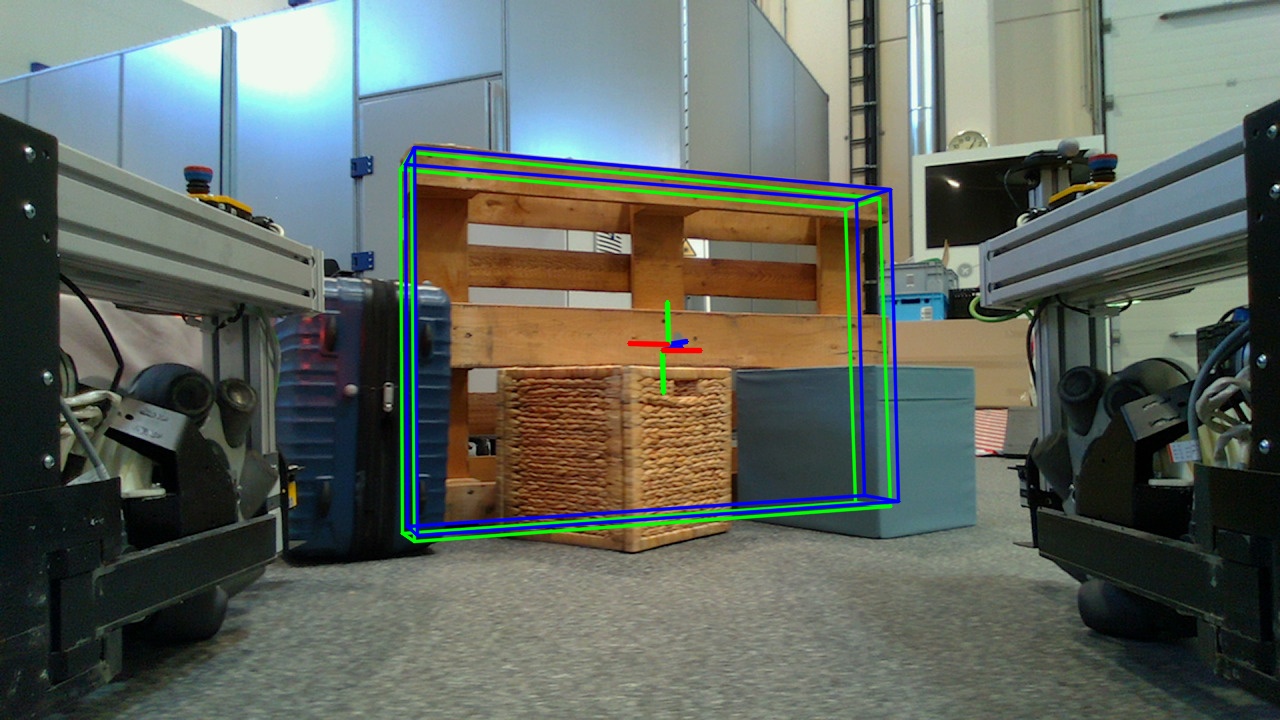}
    \includegraphics[height=2.4cm]{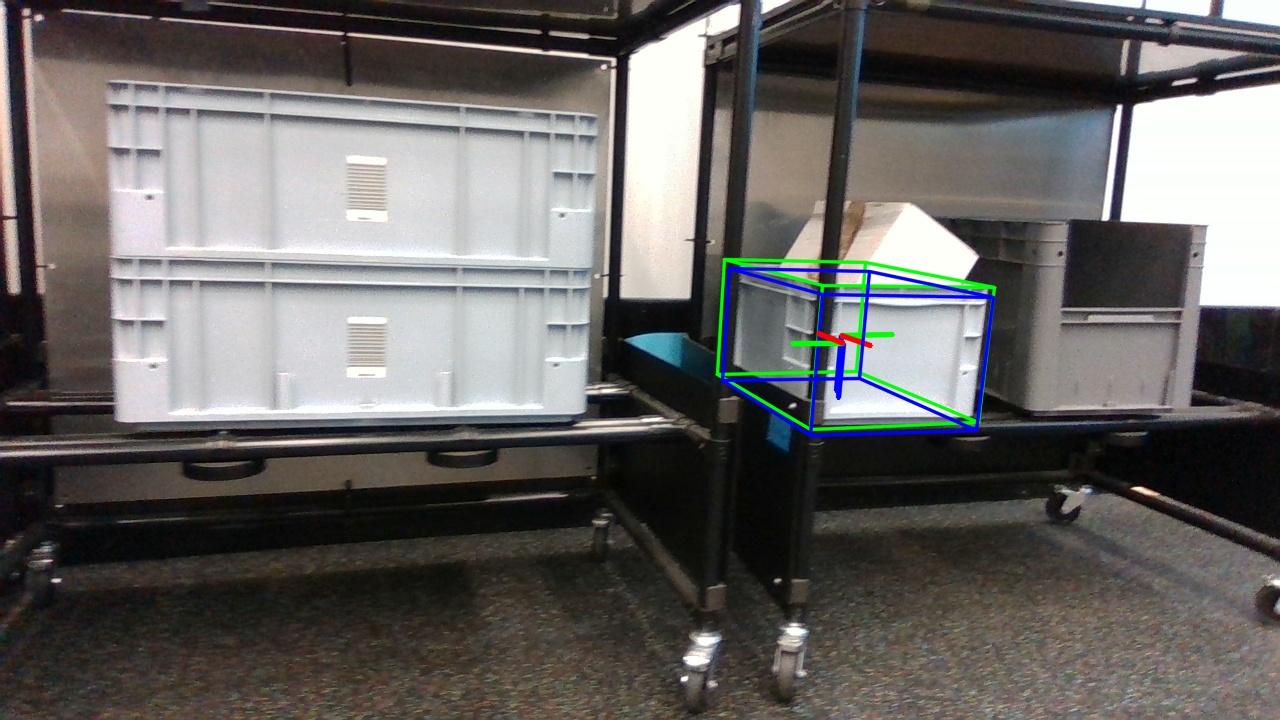}
    \\
    \caption{Quantitative results for GT-masks + FoundationPose. Ground truth (GT) is shown in green, and FoundationPose results are shown in blue.}
    \label{fig:eval_vis}
\end{figure*}

%% file: figures/fig_failure_cases.tex
\begin{figure*}
    \centering
    \includegraphics[height=2.8cm]{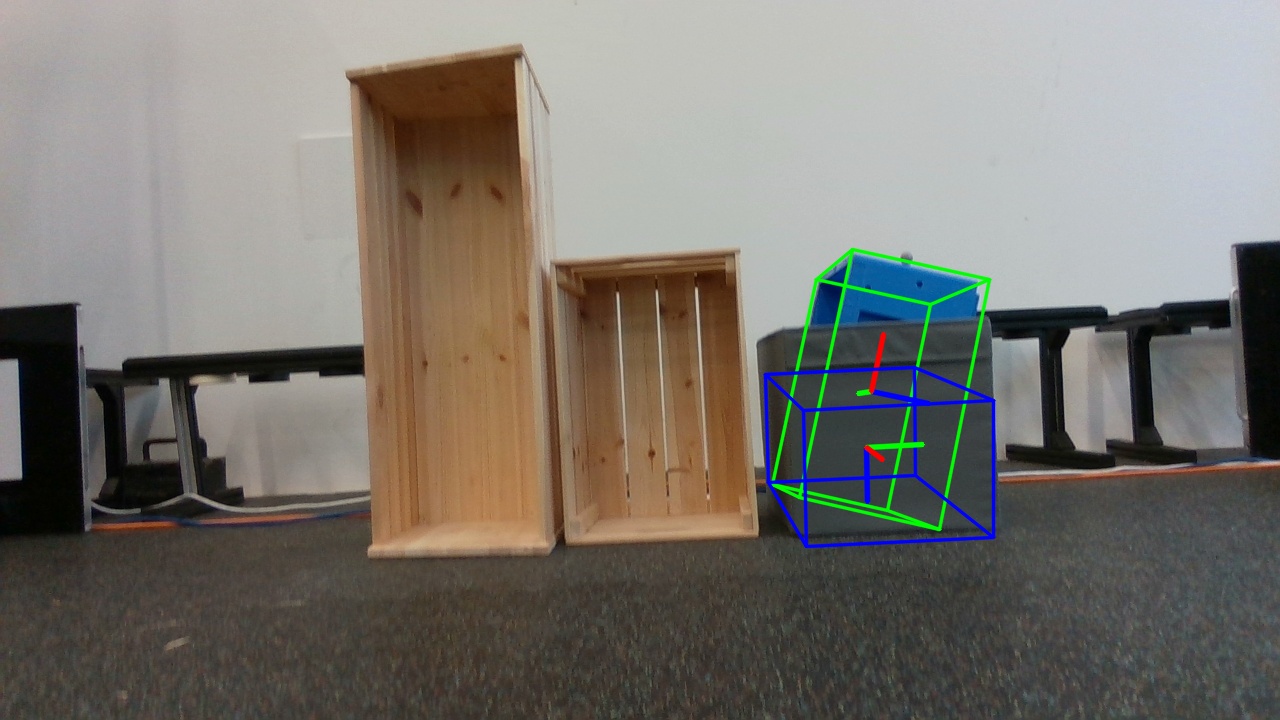}
    \includegraphics[height=2.8cm, trim=5 5 5 5, clip]{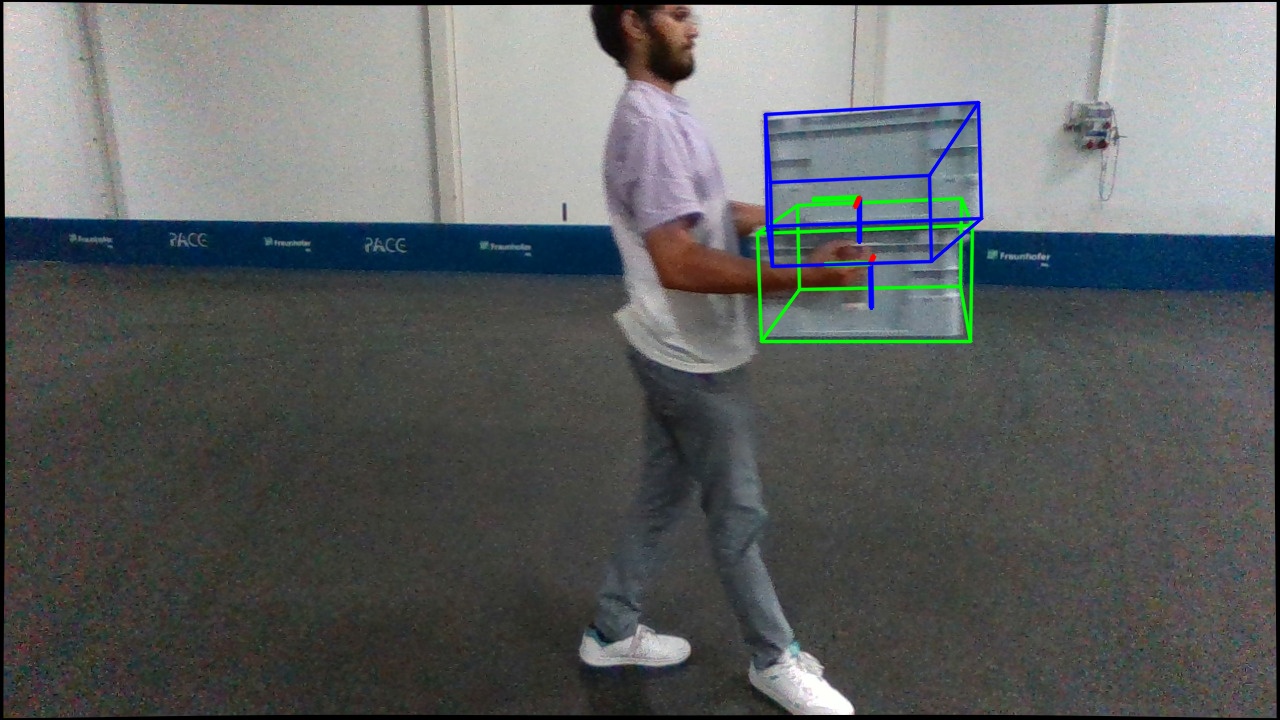}
    \includegraphics[height=2.8cm]{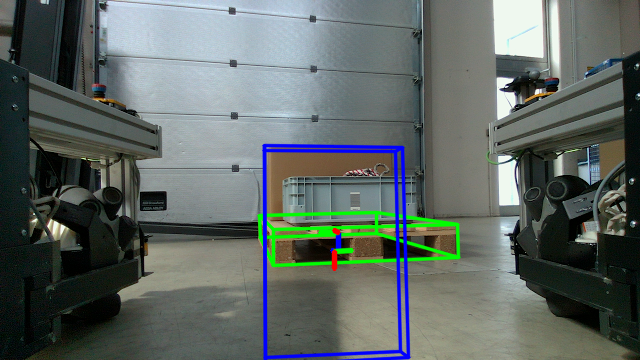}
    \\
    \caption{Challenging cases in evaluation with GT-masks + FoundationPose. Ground truth (GT) is shown in green, and FoundationPose predictions are shown in blue.}
    \label{fig:failure_cases}
\end{figure*}

%% file: sec/5_conclusion_future_work.tex
\section{Conclusion}

In this paper, we introduced MR6D, a novel dataset for 6D pose estimation in mobile robotics. Unlike existing datasets that primarily focus on small household objects manipulated by robotic arms, MR6D targets real-world challenges encountered by mobile robots, including long-range detection, extreme viewing angles, and occlusions. Comprising 92 scenes and 16 unique objects, MR6D provides a comprehensive benchmark for evaluating pose estimation models in industrial and logistics environments. Our evaluation showed that while 6D pose estimation models generalize to some extent to unseen objects, 2D segmentation of unseen objects remains a hurdle in fully unseen pipelines. Quantitative analysis indicates that self-occlusion, poor initialization due to depth values for FoundationPose, and inaccuracies in the 2D segmentation method are the main contributors to incorrect detections.

We believe that extending the current metrics used in the BOP benchmark could make it more suitable for mobile robots. In particular, we suggest extending current metrics to penalize pose errors based on tracking distance, assigning higher weights to closer objects and lower weights to those farther away. For future work, improving detection accuracy in unseen pipelines could benefit from background removal and monocular depth estimation to enhance category-agnostic segmentation. While Segment Anything is widely adopted, it often over-segments objects into smaller parts, which is detrimental to 6D pose estimation. A more suitable approach may be entity-level segmentation, where each object that cannot be physically divided is treated as a single instance. This better preserves object integrity and provides more reliable input for downstream pose estimation.